\documentclass[10pt,twocolumn,letterpaper]{article}

\usepackage[pagenumbers]{cvpr} 

\usepackage{graphicx}
\usepackage{amsmath}
\usepackage{amssymb}
\usepackage{booktabs}
\usepackage{mathtools}
\usepackage{sg-macros}
\usepackage{multirow}
\usepackage[table,xcdraw,dvipsnames]{xcolor}
\usepackage{algorithm}
\usepackage{algpseudocode}
\usepackage[accsupp]{axessibility} 

\algrenewcommand\algorithmicprocedure{\textbf{function}}

\usepackage[pagebackref,breaklinks,colorlinks]{hyperref}

\usepackage[capitalize]{cleveref}
\crefname{section}{Sec.}{Secs.}
\Crefname{section}{Section}{Sections}
\Crefname{table}{Table}{Tables}
\crefname{table}{Tab.}{Tabs.}

\def\cvprPaperID{10760} 
\def\confName{CVPR}
\def\confYear{2024}

\begin{document}

\title{Temporally Consistent Unbalanced Optimal Transport \\
    for Unsupervised Action Segmentation}

\author{Ming Xu\\
Australian National University\\
{\tt\small mingda.xu@anu.edu.au}
\and
Stephen Gould\\
Australian National University\\
{\tt\small stephen.gould@anu.edu.au}
}
\maketitle

\newcommand{\methodname}{\text{ASOT}}
\begin{abstract}
   We propose a novel approach to the action segmentation task for long, untrimmed videos, based on solving an optimal transport problem. By encoding a temporal consistency prior into a Gromov-Wasserstein problem, we are able to decode a temporally consistent segmentation from a noisy affinity/matching cost matrix between video frames and action classes. Unlike previous approaches, our method does not require knowing the action order for a video to attain temporal consistency. Furthermore, our resulting (fused) Gromov-Wasserstein problem can be efficiently solved on GPUs using a few iterations of projected mirror descent. We demonstrate the effectiveness of our method in an unsupervised learning setting, where our method is used to generate pseudo-labels for self-training. We evaluate our segmentation approach and unsupervised learning pipeline on the Breakfast, 50-Salads, YouTube Instructions and Desktop Assembly datasets, yielding state-of-the-art results for the unsupervised video action segmentation task.
\end{abstract}

\section{Introduction}
\label{sec:intro}

While action recognition is a well-studied topic in the area of video understanding, datasets are typically comprised of short videos arising from distinctive action categories with tightly cropped temporal boundaries~\cite{kay2017kinetics, KarpathyCVPR14}. In this paper, we study the less-explored setting of segmenting long, untrimmed videos of multi-stage activities, where each video contains multiple actions. A fine grained temporal understanding of videos is required for this task, which can be attained in a supervised learning setting given dense, per-frame annotations. Developing techniques to address this task in an unsupervised manner will allow large video collections to be utilized for learning without requiring expensive frame-level annotations.

The action segmentation task can be formulated as a dense classification problem, where each frame within a video is assigned to an action class. Specific to this task, however, is that individual actions occupy temporal blocks of frames. We call this phenomena the \textit{temporal consistency} property. Current approaches in the unsupervised setting do not provide temporally consistent predictions ``out-of-the-box", necessitating \textit{post-processing} methods based on hidden Markov models (HMMs)~\cite{Kukleva_2019_CVPR, ude, VidalMata_2021_WACV, Li_2021_CVPR, Kumar_2022_CVPR, tran2023permutationaware, tempseg_survey} to decode a segmentation from the learned representations. A critical assumption for these HMM methods is that the action order present within a video is known. 

\begin{figure}[t!]
    \centering
    \includegraphics[width=0.45\textwidth]{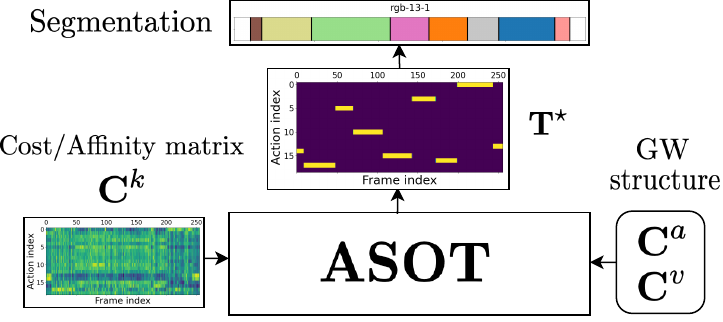}
    \caption{High-level overview of our action segmentation method, \methodname{}. Given a (noisy) cost/affinity matrix between video frames and actions, \methodname{} solves an optimal transport (OT) problem to yield temporally consistent segmentations.}
    \label{fig:system_asot}
\end{figure}

Furthermore, successful approaches to unsupervised learning for image data based on joint representation learning and clustering~\cite{swav_neurips_20, asanoself20} have been adapted to the action segmentation task~\cite{Kumar_2022_CVPR, tran2023permutationaware}, yielding state-of-the-art results. These methods use (regularized) optimal transport (OT)~\cite{cuturi_lightspeed} to generate pseudo-labels for self-training, and jointly learn a video encoder and a set of action class embeddings. However, the OT formulation used in~\cite{Kumar_2022_CVPR, tran2023permutationaware} does not account for temporal consistency, which is inherent to the action segmentation task. In addition, the balanced assignment assumption is imposed, which yields pseudo-labels that are uniformly distributed among action classes. We argue this is unreasonable for action segmentation since datasets used~\cite{breakfast, Alayrac_2016_CVPR, salads} exhibit long-tailed class distributions~\cite{tempseg_survey}.

Motivated by these observations, we propose Action Segmentation Optimal Transport (\methodname{}), a novel method for decoding a temporally consistent segmentation from a noisy affinity/matching cost matrix between video frames and action classes. \methodname{} involves solving a fused, unbalanced Gromov-Wasserstein (GW) optimal transport (OT) problem, which fuses visual information from the cost matrix with a structure-aware GW component that encourages temporal consistency. Unbalanced OT allows for only a subset of actions to be represented within a video. Fig.~\ref{fig:system_asot} illustrates the mechanics of \methodname{}.


Finally, we show that \methodname{} is generally applicable as a post-processing step for action segmentation pipelines. Unlike prior HMM-based approaches, \methodname{} handles order variations and repeated actions without knowing the action ordering. While we primarily evaluate \methodname{} in an unsupervised learning setting, we also show how \methodname{} is effective for post-processing of supervised methods. We thoroughly evaluate \methodname{} on the unsupervised action segmentation task, yielding state-of-the-art results. We provide an implementation of \methodname{} with training and evaluation code\footnote{\url{https://github.com/mingu6/action_seg_ot}}.

\section{Related Work}\label{sec:related}

\paragraph{Fully Supervised Action Segmentation.} 
Recent approaches for the action segmentation task in the fully-supervised setting have focused on developing appropriate model architectures. In particular, the temporal convolutional network (TCN)~\cite{Lea_2017_CVPR, Lei_2018_CVPR} is widely used in the literature since it handles long-term dependencies through a temporal convolution layer. MS-TCN~\cite{Farha_2019_CVPR, mstcnplus} uses a multi-scale TCN so videos can be processed at a high temporal resolution. Recently, transformer style architectures are also being investigated~\cite{chinayi_ASformer, uvast2022ECCV, azieremstct22}, yielding promising results. Finally, some works propose additional segmentation refinement blocks on top of existing architectures~\cite{isikawa_wacv, Ahn_2021_ICCV, Chen_2020_CVPR}. Ad hoc smoothness losses are commonly included during training to encourage temporal consistency~\cite{Farha_2019_CVPR, Chen_2020_CVPR, chinayi_ASformer, Guan_2021_ICCV}. We show how \methodname{} can be used as a post-processing step to further improve the performance of MS-TCN~\cite{mstcnplus}.


\paragraph{Unsupervised Action Segmentation.}
In the unsupervised setting, learning by solving a proxy task is a common approach. Kukleva~\etal~\cite{Kukleva_2019_CVPR} proposed CTE, which uses timestamp prediction as a proxy task. The output from intermediate network layers is used for the frame embeddings. UDE~\cite{ude} and VTE~\cite{VidalMata_2021_WACV} extend CTE with a discriminative embedding loss and visual reconstruction loss. These methods have two stages; representations are learned through the proxy task, and a clustering algorithm is subsequently used to recover actions. ASAL~\cite{Li_2021_CVPR} alternate between representation learning and HMM parameter estimation using a generalized EM approach. Notably, TOT~\cite{Kumar_2022_CVPR} and UFSA~\cite{tran2023permutationaware} are similar to our approach, performing joint representation learning and clustering using temporal OT. 

However, TOT~\cite{Kumar_2022_CVPR} suffers from three limitations: 1) a balanced assignment assumption is enforced on pseudo-labels, which we argue is unreasonable for long-tailed action class distributions encountered in action segmentation datasets~\cite{tempseg_survey}, 2) temporal consistency is not addressed and 3) a single, fixed action ordering for all videos is assumed. UFSA~\cite{tran2023permutationaware} extends TOT~\cite{Kumar_2022_CVPR} by relaxing 1) and 3) through two learned modules, an action transcript prediction module and a frame to transcript alignment module. In contrast, our method addresses all three limitations of TOT, handling order variations and repeated actions, by expanding the OT formulation, with no significant increase in learnable parameters or network architecture complexity from UFSA~\cite{tran2023permutationaware}. Furthermore, TOT~\cite{Kumar_2022_CVPR} and UFSA~\cite{tran2023permutationaware} use a HMM approach to decode segmentations given a fixed (TOT) or estimated (UFSA) action ordering. In contrast, we can use \methodname{} for both pseudo-labelling and decoding.

We refer the reader to Ding~\etal~\cite{tempseg_survey} for a comprehensive survey on the temporal action segmentation task.

\paragraph{Optimal Transport for Structured Prediction.} 

Optimal transport has been used for many alignment tasks in computer vision and machine learning, with too many applications to be listed here. See Khamis et al.~\cite{khamis2023earth} for a recent survey. Recent examples in computer vision include keypoint matching across image pairs~\cite{Sarlin_2020_CVPR, liu2020semantic}, point set registration~\cite{shenrobot21} and training object detectors~\cite{De_Plaen_2023_CVPR}. GW OT in particular, has been used for structured data, \eg representation learning on graphs~\cite{xu19gw} and recently, aligning brains using fMRI scans\cite{fgw_brain}. For the first time, we use GW to exploit problem structure for a segmentation task. Our method has strong connections to post-processing methods for image segmentation problems, e.g.,~dense conditional random fields~\cite{efficient_crf_2011} and bilateral filtering~\cite{bilateralsolver}, which is worth exploring in more detail in future work.
\section{Optimal Transport on Structured Data}

Our proposed method, \methodname{}, formulates the post-processing step of the temporal action segmentation task as an optimal transport problem. Specifically, we adopt a fused, unbalanced, Gromov-Wasserstein (FUGW OT) formulation. In this section, we will provide some background on optimal transport and specifically, the FUGW problem.

\paragraph{Preliminaries.} First, let $\langle \mathbf{A}, \mathbf{B} \rangle \coloneqq \sum_{i, j} A_{ij}, B_{ij}$ for $\mathbf{A}, \mathbf{B}\in\mathbb{R}^{n\times m}$ and let $\mathbf{1}_n$ be a length $n$ vector of ones. Denote a dataset comprised of $B$ videos as $\mathcal{D}\coloneqq \{\mathcal{V}^b\}_{b=1}^B$. Each video is assumed to be comprised of $N$ frames, denoted $\mathcal{V}^b \coloneqq \{\mathcal{I}_{1}^{b}, \dots, \mathcal{I}_{N}^{b}\}$. Furthermore, let $\mathbf{X}^b \coloneqq f_\theta(\mathcal{V}^b)\in\mathbb{R}^{N\times D}$ denote frame-level embeddings for a single video, extracted using a deep network parameterized by $\theta$. Let (learnable) cluster centroids, which represent actions, be denoted $\mathbf{A}\coloneqq \left[ \mathbf{a}_1, \dots, \mathbf{a}_K \right] \in\mathbb{R}^{D\times K}$. Let $[n]\coloneqq \{1,\dots,n\}$ denote a discrete set with $n$ elements, and finally, let $\Delta_K\subset \mathbb{R}^K$ be the $K-1$ dimensional probability simplex and $\Delta_K^N\subset \mathbb{R}^{K\times N}$ denote the Cartesian product space comprised of $N$ such simplexes.

\subsection{Kantorovich Optimal Transport}\label{ssec:kot}

The classic optimal transport formulation is commonly known as the Kantorovich (KOT) formulation~\cite{thorpeot}, and we focus on the discrete setting in this paper. Given histograms $\mathbf{p}\in \Delta_n$ and $\mathbf{q} \in \Delta_m$, and a ground cost $\mathbf{C}^k\in\mathbb{R}^{n\times m}_+$, the KOT problem solves for the minimum cost coupling $\mathbf{T}^\star$ between $\mathbf{p}$ and $\mathbf{q}$. Specifically, the problem is given by
\begin{equation}\label{eq:ot_prob}
    \begin{array}{cc}
       \underset{\mathbf{T}\in \mathcal{T}_{\mathbf{p}, \mathbf{q}}}{\text{minimize}} & \mathcal{F}_{\text{KOT}}(\mathbf{C}^k, \mathbf{T}) \coloneqq \langle \mathbf{C}^k, \mathbf{T}\rangle,
    \end{array}
\end{equation}
where $\mathcal{T}_{\mathbf{p}, \mathbf{q}} \coloneqq \{\mathbf{T}\in\mathbb{R}_+^{n\times m} \mid \mathbf{T} \ones_m = \mathbf{p}, \mathbf{T}^\top \ones_n = \mathbf{q} \}$ is commonly referred to as the \textit{transportation polytope}. The coupling $\mathbf{T}$ can be interpreted as a soft \textit{assignment} between elements in the support of $\mathbf{p}$ and $\mathbf{q}$, namely, two discrete sets denoted $[n]$ and $[m]$. In the context of action segmentation, coupling $\mathbf{T}\in\mathbb{R}^{N\times K}$ can be interpreted as an assignment between frames and actions.

\subsection{Gromov-Wasserstein Optimal Transport}\label{ssec:gwot}

Gromov-Wasserstein (GW) optimal transport is an extension to the Kantorovich formulation, and is used for comparing histograms defined over incomparable spaces. Concretely, let $(\mathbf{C}^v, \mathbf{p})\in\mathbb{R}^{n\times n}\times \Delta_n$ and $(\mathbf{C}^a, \mathbf{q})\in\mathbb{R}^{m\times m}\times \Delta_m$ be two metric-measure pairs. The distance matrices, $\mathbf{C}^v$ (resp.~$\mathbf{C}^a)$ fully describe a metric defined over supports $[n]$ (resp.~$[m]$). In the GW setting, there is no metric/cost defined \textit{between} $[n]$ and $[m]$. The GW OT problem~\cite{icml_16_peyre_gw} replaces the KOT objective function in Eq.~\ref{eq:ot_prob} with

\begin{equation}\label{eq:gwot_obj}
    \mathcal{F}_{\text{GW}}(\mathbf{C}^v, \mathbf{C}^a, \mathbf{T}) \coloneqq \sum_{\substack{i,k \in [n]\\j,l \in [m]}} L(C_{ik}^v, C_{jl}^a) T_{ij}T_{kl},
\end{equation}
where $L:\mathbb{R}\times \mathbb{R} \rightarrow \mathbb{R}$ is a cost function penalizing deviations \textit{between distance matrix elements}. GW OT is commonly used to compare structured objects,~\eg graphs~\cite{icml_16_peyre_gw}. In our work, we use this GW formulation to encode \textit{structural priors} over the transport map desirable for the video segmentation task (\ie temporal consistency). This is achieved by setting $\mathbf{C}^v$ and $\mathbf{C}^a$ in a particular way, which will be described in detail in Sec.~\ref{ssec:costvideo}. 

\subsection{Fused GW Optimal Transport}\label{ssec:fgw}

The fused Gromov-Wasserstein (FGW) problem~\cite{vayer_fgw_algo, li_23_jcgs, fgw_brain, pmlr-v97-titouan19a} combines the KOT and GW OT formulations into a single optimization problem. The FGW formulation is useful when both a ground cost and a structural prior is available. FGW OT has been used previously in machine learning for graph classification and clustering~\cite{vayer_fgw_algo} and aligning fMRI data~\cite{fgw_brain}. Given parameter $\alpha \in [0, 1]$, and letting $\mathbf{C}\coloneqq \{\mathbf{C}^k, \mathbf{C}^v, \mathbf{C}^a\}$, the FGW objective is given by 
\begin{equation}\label{eq:fgwot_obj}
    \mathcal{F}_{\text{FGW}}(\mathbf{C}, \mathbf{T}) \coloneqq \alpha\mathcal{F}_{\text{GW}}(\mathbf{C}^v, \mathbf{C}^a, \mathbf{T}) + (1-\alpha) \mathcal{F}_{\text{KOT}}(\mathbf{C}^k, \mathbf{T}).
\end{equation}
For the temporal segmentation problem, the KOT component encodes visual similarity between the cluster/action representations and video frame embeddings. The GW component on the other hand, encodes desirable structural properties for resulting segmentations, \ie temporal consistency. We will describe how parameters $\mathbf{C}$ are determined for action segmentation in Sec.~\ref{sec:tempseg}.

\subsection{Unbalanced Optimal Transport}\label{ssec:unbalancedot}

Recently, there has been interest in \textit{unbalanced} transport problems~\cite{fgw_brain, ubgw21, chizat_scaling_ub}, where the constraint for $\mathbf{T}$ to lie in the transportation polytope $\mathcal{T}_{\mathbf{p}, \mathbf{q}}$ (often referred to as the \textit{balanced assignment property}) is relaxed. This can be achieved by replacing the marginal constraints on $\mathbf{T}$ with penalty terms in the objective function. Concretely, in this work, we will solve the optimization problem given by
\begin{equation}\label{eq:fugw_prob}
    \begin{array}{cc}
       \underset{\mathbf{T}\in\mathcal{T}_{\mathbf{p}}}{\text{minimize}} & \mathcal{F}_{\text{FGW}}(\mathbf{C}, \mathbf{T}) +  \lambda \text{D}_{\text{KL}}(\mathbf{T}^\top \ones_n \| \mathbf{q}),
    \end{array}
\end{equation}
where $\text{D}_\text{KL}(\mathbf{a} \| \mathbf{b})\coloneqq \sum_i a_i \log (a_i / b_i)$ is the Kullback-Leibler divergence and $\mathcal{T}_{\mathbf{p}} \coloneqq \{\mathbf{T}\in\mathbb{R}_+^{n\times m} \mid \mathbf{T} \ones_K = \mathbf{p}\}$ is a \textit{partial polytope constraint} over the row-sum marginal of $\mathbf{T}$. A penalty, weighted by $\lambda > 0$, is applied between the column-sum marginal of $\mathbf{T}$ to $\mathbf{q}$. We will explain in Sec.~\ref{ssec:unbalancedvideo} why the unbalanced formulation is important for the temporal segmentation task. 

\section{Action Segmentation Optimal Transport }\label{sec:tempseg}

In this section, we describe our proposed post-processing approach \methodname{}, for the action segmentation problem. First, we explain how to extract a temporally consistent segmentation from a (noisy) cost matrix between video frames and actions using optimal transport. These cost matrices are easily computed using learned video frame and action embeddings. Second, we introduce the fused, unbalanced Gromov-Wasserstein OT problem underlying \methodname{}. The importance of the unbalanced and GW components for the action segmentation problem will be described in detail. Finally, we discuss including entropy regularization, which allows us to develop a fast, iterative solution based on scaling algorithms~\cite{chizat_scaling_ub, ubgw21, fgw_brain, icml_16_peyre_gw, cuturi_lightspeed}.

\subsection{Optimal Transport for Action Segmentation}

Let $\mathbf{p} = \frac{1}{N}\ones_N$ and $\mathbf{q} = \frac{1}{K}\ones_K$ be histograms over the set of $N$ video frames and $K$ actions, denoted $[N]$ and $[K]$, respectively. An (optimal) coupling $\mathbf{T}^\star\in\mathbb{R}^{N\times K}$ between $[N]$ and $[K]$ can be interpreted as a (soft) \textit{assignment} between video frames and actions. For frame $i$, we can predict the corresponding action using $j^\star = \argmax_j\mathbf{T}_{ij}^\star$. Let $\mathbf{C}$ be the set of cost matrices for the KOT and GW sub-problems, which will be defined in Sec.~\ref{ssec:costvideo}. We solve the optimization problem in Eq.~\ref{eq:fugw_prob} for the coupling $\mathbf{T}^\star$.

\subsection{Objective Function Formulation}\label{ssec:costvideo}

\paragraph{Visual Information.} 

Given frame embeddings $\mathbf{X}$ and action embeddings $\mathbf{A}$, we can derive the matching cost matrix as $C^k_{ij} \coloneqq 1 - \frac{\mathbf{x}_i^\top \mathbf{a}_j}{\|\mathbf{x}_i\|_2\|\mathbf{a}_j\|_2}$. This defines the KOT component of Eq.~\ref{eq:fugw_prob}, and incorporates visual information from the videos through the frame encoder. However, the KOT component alone does not provide temporal consistency.

\paragraph{Structural Priors.} 
The GW component encourages temporal consistency over $\mathbf{T}$. Define the GW cost function as $L(a, b)\coloneqq ab$, radius parameter $r \in [0, 1]$ and furthermore, let $\mathbf{C}^v\in\mathbb{R}^{N\times N}$ and $\mathbf{C}^a\in\mathbb{R}^{K\times K}$ be costs over video frames and action categories, respectively, given by
\begin{equation}\label{eq:cv_def}
    \mathbf{C}^v_{ik} \coloneqq 
    \begin{cases}
        1 / r & 1 \leq |i - k| \leq Nr \\
        0 & \text{o/w}
    \end{cases} , \,\, \mathbf{C}^a_{jl} \coloneqq 
    \begin{cases}
        0 & j = l \\
        1 & \text{o/w}
    \end{cases}.
\end{equation}

This GW formulation penalizes couplings with low temporal consistency. Specifically, assigning temporally adjacent frames (within a radius of $Nr$ frames) to different actions incurs a cost in the GW objective in Eq.~\ref{eq:gwot_obj}. To see this, let $T_{ij}$ and $T_{kl}$ represent two assignment probabilities with $|i-k| \leq Nr$, $i\neq k$ (adjacent) and $j \neq l$ (different actions). A cost of $T_{ij}T_{kl}$ is incurred in Eq.~\ref{eq:gwot_obj}, since $L(C_{ik}^\mathbf{p}, C_{jl}^\mathbf{q}) = 1$. On the other hand, action changes outside of a temporal interval ($|i-k| > Nr$) incur no cost nor does mapping adjacent frames to the same action ($j=l$). Fig.~\ref{fig:asot_example} and~\ref{fig:nogw_example} illustrates the effect of the GW component.

Finally, note that we can compute Eq.~\ref{eq:gwot_obj} efficiently as $\mathcal{F}_{\text{GW}}(\mathbf{C}^v, \mathbf{C}^a, \mathbf{T}) = \langle \mathbf{C}^v \mathbf{T} \mathbf{C}^a , \mathbf{T} \rangle$ since the loss $L(a, b) = ab$ can be factorized, following Peyr\'{e} et al.~\cite{icml_16_peyre_gw}.

\subsection{Importance of Unbalanced Transport}\label{ssec:unbalancedvideo}

\begin{figure}[t!]
     \centering
     \begin{subfigure}[b]{0.21\textwidth}
         \centering
         \includegraphics[width=\textwidth]{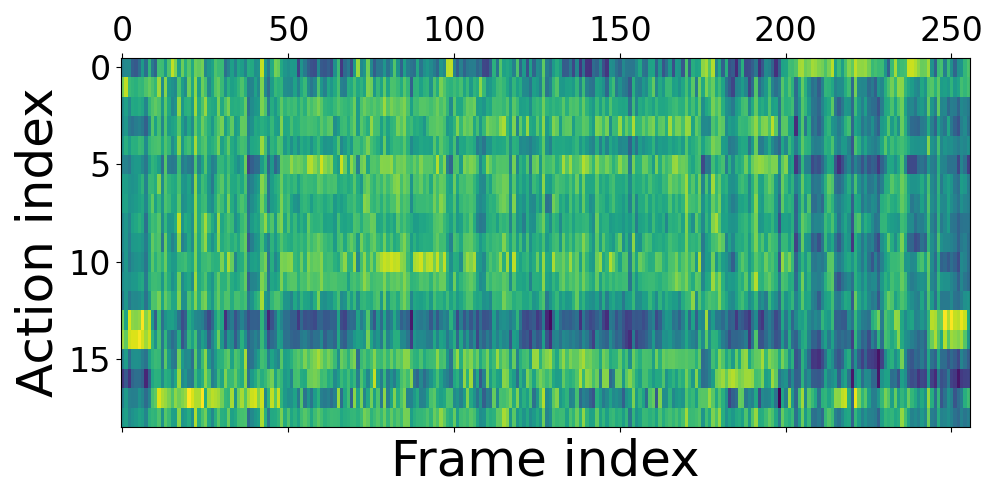}
         \caption{Affinity matrix $1 - \mathbf{C}^k$.}
         \label{fig:raw_aff_example}
     \end{subfigure}
     \begin{subfigure}[b]{0.21\textwidth}
         \centering
         \includegraphics[width=\textwidth]{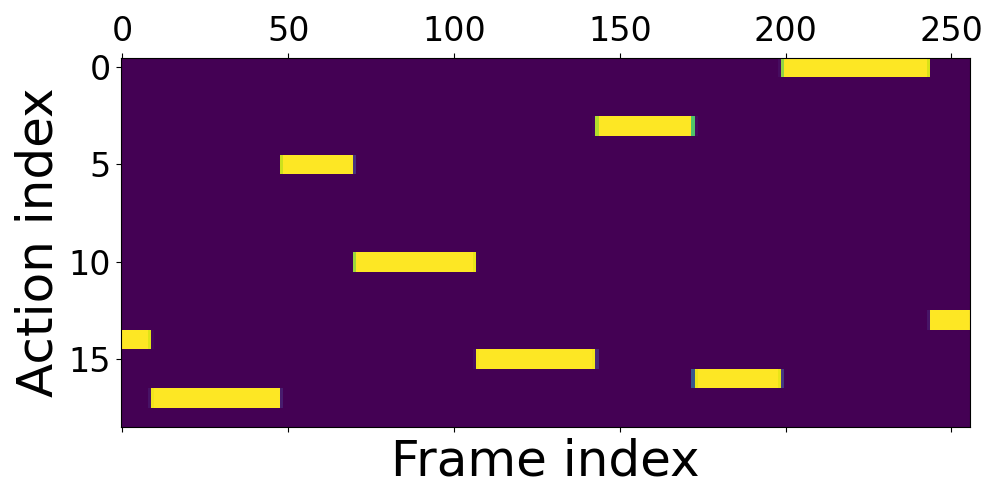}
         \caption{\methodname{} solution.}
         \label{fig:asot_example}
     \end{subfigure}
     
     \begin{subfigure}[b]{0.21\textwidth}
         \centering
         \includegraphics[width=\textwidth]{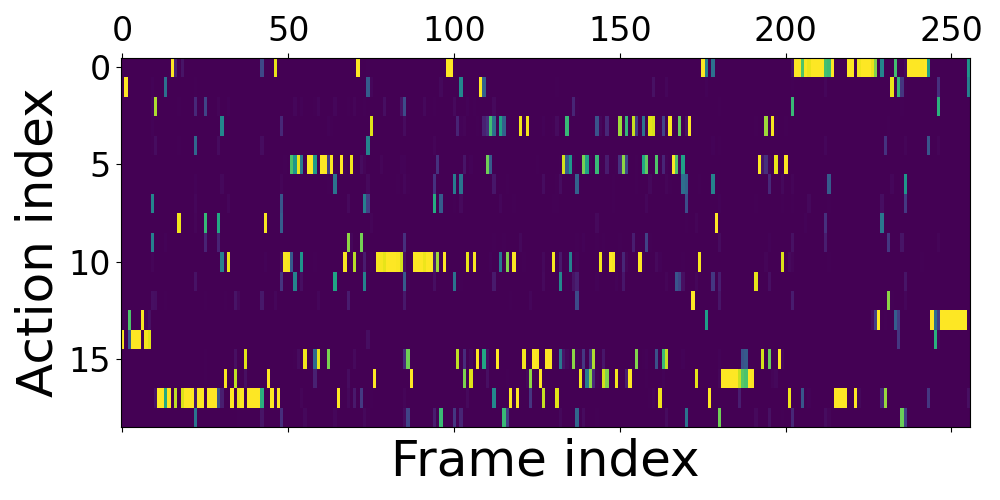}
         \caption{\methodname{} w/no GW ($\alpha = 0$).}
         \label{fig:nogw_example}
     \end{subfigure}
     \begin{subfigure}[b]{0.21\textwidth}
         \centering
         \includegraphics[width=\textwidth]{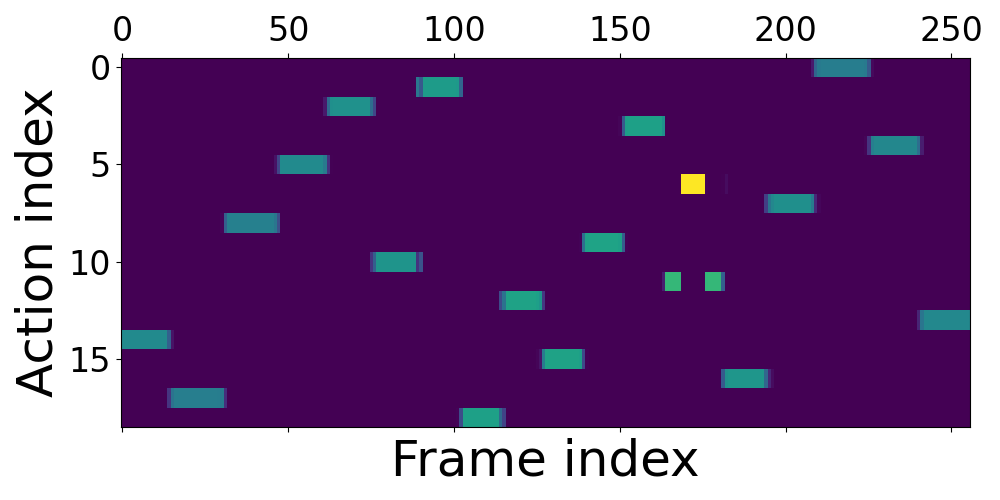}
         \caption{\methodname{} w/out unbalanced OT.}
         \label{fig:balanced_example}
     \end{subfigure}
    \caption{A raw frame/action affinity matrix in a) is decoded using \methodname{} into a temporally consistent segmentation in b). Removing GW from b) destroys temporal consistency, shown in c). Forcing a balanced assignment to actions in b) yields temporally consistent, but unintuitive results, shown in d).}
    \label{fig:qual_fugw}
\end{figure}

For the temporal action segmentation task, a balanced assignment necessitates that each video frame is assigned to an action and furthermore, that each action is represented equally across the frames of a video. While assigning each video frame to an action is required for the segmentation task, not every action may occur within a video. The partial polytope constraint in Sec.~\ref{ssec:unbalancedot}, \ie $\mathbf{T}\in \mathcal{T}_{\mathbf{p}}$, ensures that every frame is assigned to an action. However, we do not force every action to be evenly represented over video frames. Instead, we regularize the distribution of actions observed within a video towards uniformity using a KL-divergence penalty weighted by $\lambda > 0$. Setting lower values for $\lambda$ allows \methodname{} to find a segmentation that faithfully replicate the learned embeddings, whereas a high $\lambda$ emphasises that the segmentation is balanced across actions. Fig.~\ref{fig:asot_example} and~\ref{fig:balanced_example} illustrates balanced and unbalanced OT.

\subsection{Fast Numerical Solver for \methodname{}}

We add an entropy regularization term $-\epsilon H(\mathbf{T})$, where $H(\mathbf{T}) \coloneqq -\sum_{i,j} T_{ij}\log T_{ij}$  and $\epsilon > 0$, into the objective function in Eq.~\ref{eq:fugw_prob}. We can solve the FUGW problem using projected mirror descent similar to~\cite{icml_16_peyre_gw}, which is amenable for computation on GPUs. Our solver typically converges within 25 iterations, with each iteration having time complexity $O(NK)$ after exploiting the sparsity structure of $\mathbf{C}^v$ and $\mathbf{C}^a$. We provide more details and psuedocode in the appendix. The largest video we encountered (in the 50Salads~\cite{salads} dataset) had $N=16k$ frames and $K=19$ action classes and takes 26.1ms on an Nvidia RTX 4090 GPU.

\section{Unsupervised Learning Pipeline}\label{sec:replearning}

\begin{figure}[t!]
    \centering
    \includegraphics[width=0.40\textwidth]{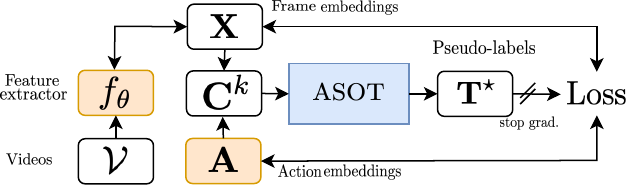}
    \caption{The unsupervised training pipeline. Orange are learnable parameters, and arrows indicate computation/gradient flow.}
    \label{fig:train}
\end{figure}

In this section, we will describe our simple representation learning pipeline for unsupervised action segmentation. Core to our method is a self-training approach; we use \methodname{} described in Sec.~\ref{sec:tempseg} to generate pseudo-labels, which are then used to train a video frame encoder. Fig.~\ref{fig:train} illustrates the representation learning pipeline.

\paragraph{Learning Problem.} 
Our unsupervised problem involves learning the parameters $\theta$ of a video frame encoder by minimizing the cross-entropy (CE) loss between frame/action embedding similarities and (soft) pseudo-labels computed by \methodname{}. Concretely, let $\mathbf{P}^b\in\Delta_K^N$ be normalized similarities for batch element $b$, defined element-wise as
\begin{equation}\label{eq:clus-emb-sims}
    P^b_{ij} \coloneqq \frac{\exp(\mathbf{X}^b \mathbf{A}^\top / \tau)_{ij}}{\sum_l\exp(\mathbf{X}^b \mathbf{A}^\top / \tau)_{il}},
\end{equation}
where $\tau > 0$ is a temperature scaling parameter. Next, soft-pseudo labels $\mathbf{T}^b\in \mathbb{R}^{K \times N}$ are derived from the solution to the optimization problem in Eq.~\ref{eq:fugw_prob}, parameterized using a KOT cost matrix given by $\hat{\mathbf{C}}^k = \mathbf{C}^k + \rho \mathbf{R}$ where $\rho \geq 0$. 

Here $\mathbf{R}$ is a \textit{temporal prior}, introduced in~\cite{Kumar_2022_CVPR}, defined element-wise as $R_{ij} = \left| i / N - j / K \right|$. $\mathbf{R}$ regularizes the coupling towards a banded diagonal and encourages a canonical ordering of actions across videos. We find setting $\rho > 0$ encourages correspondences between adjacent frames \textit{across videos}, improving clustering performance.

Our representation learning loss is given by
\begin{equation}\label{eq:crossentropy}
    \mathcal{L}_{\text{train}}(\theta) = -\frac{1}{B}\sum_{b=1}^B \sum_{i=1}^N\sum_{j=1}^k T^b_{ij} \log P^b_{ij}.
\end{equation}
Same as in prior works~\cite{swav_neurips_20, Kumar_2022_CVPR}, we apply a stop-gradient through pseudo-labels $\mathbf{T}^b$. The learnable parameters include the feature extractor $\theta$ as well as action embeddings $\mathbf{A}$. We parameterize $f_\theta$ as a multi-layer perceptron (MLP) applied per frame, consistent with prior works~\cite{Kukleva_2019_CVPR, Li_2021_CVPR, Kumar_2022_CVPR}. 

\section{Experimental Setup}\label{sec:exper}

\paragraph{Implementation Details.} 
The encoder MLP has one hidden layer and we use the Adam optimizer~\cite{kingma:adam} with a learning rate of $10^{-3}$ and weight decay of $10^{-4}$. We use $k$-means to initialize the action embeddings. During training, we sample 256 frames randomly from uniformly spaced bins within each video, similar to~\cite{Kumar_2022_CVPR}. A full description of hyperparameter settings is provided in the appendix. We set $K$ to be equal to the ground truth number of actions per activity category/dataset, consistent with prior works~\cite{Kukleva_2019_CVPR, Li_2021_CVPR, Kumar_2022_CVPR, tran2023permutationaware}.

\paragraph{Datasets.} 
We briefly describe the four video datasets we use to evaluate our method. All unsupervised methods are trained and evaluated on the same set of videos. While videos are subsampled during training for our method, we use full videos during testing. For Breakfast and YouTube Instructions, we train and evaluate our method per activity category and aggregate results.

\begin{itemize}
    \item The Breakfast (BF) dataset~\cite{breakfast} consists of approx. 1,700 videos, where each video captures a subject preparing a breakfast item. The dataset is split across activity categories, where each activity category corresponds to a particular item (\eg scrambled egg, juice). Within each video, multiple actions are observed (\eg crack egg, pour flour). Breakfast contains 10 activity categories with 48 actions across all activities, and videos range from a few seconds to several minutes. We use pre-computed Fisher vector features from images, consistent with prior works~\cite{Kukleva_2019_CVPR, Li_2021_CVPR, Kumar_2022_CVPR}.
    \item The YouTube Instructions (YTI) dataset~\cite{Alayrac_2016_CVPR} includes 150 instructional videos belonging to 5 activity categories. The average video lasts approx.\ 2 minutes, and this dataset also has a large number of background frames. We use pre-computed image features provided by~\cite{Alayrac_2016_CVPR}, consistent with prior works~\cite{Kukleva_2019_CVPR, Li_2021_CVPR, Kumar_2022_CVPR}.
    \item The 50 Salads (FS) dataset~\cite{salads} contains 50 videos of actors performing a cooking activity, totalling 4.5 hours in length. Consistent with prior works, we report results at two action granularity levels, \ie \textit{Mid} with 19 action classes and \textit{Eval} with 12 action classes. The Eval level aggregates some actions in the Mid level into a single action. We use pre-computed features which were used in prior works~\cite{Kukleva_2019_CVPR, Li_2021_CVPR, Kumar_2022_CVPR}.
    \item The Desktop Assembly (DA) dataset~\cite{Kumar_2022_CVPR} includes 76 videos of actors performing an assembly activity. The activity comprises 22 actions conducted in a fixed order. Each video is approx.\ 1.5 minutes long, and we use the pre-computed features provided by~\cite{Kumar_2022_CVPR}.
\end{itemize}
\begin{figure}[t!]
     \centering
     \begin{subfigure}[b]{0.35\textwidth}
         \centering
         \includegraphics[width=0.9\textwidth]{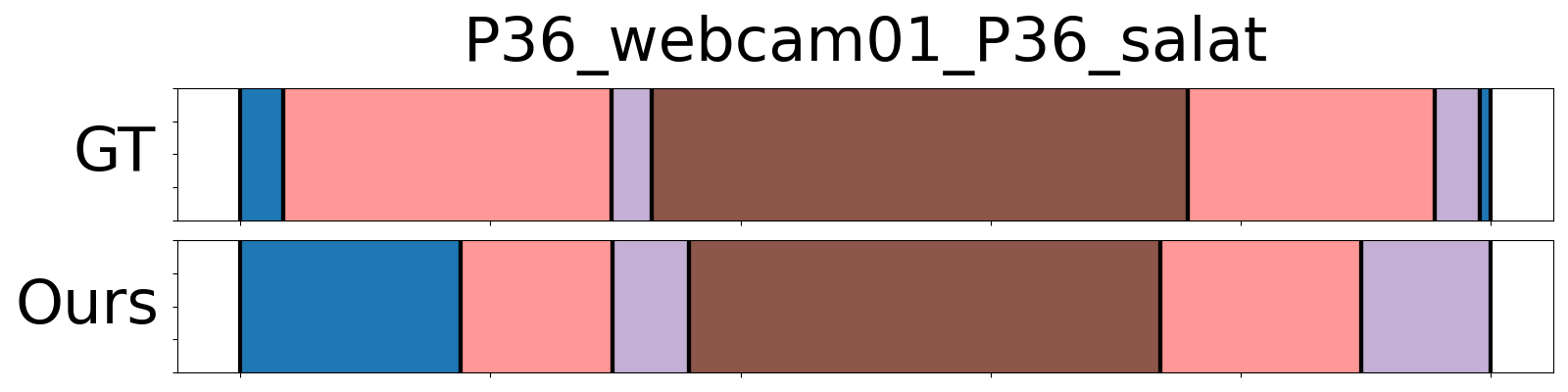}
         \includegraphics[width=0.9\textwidth]{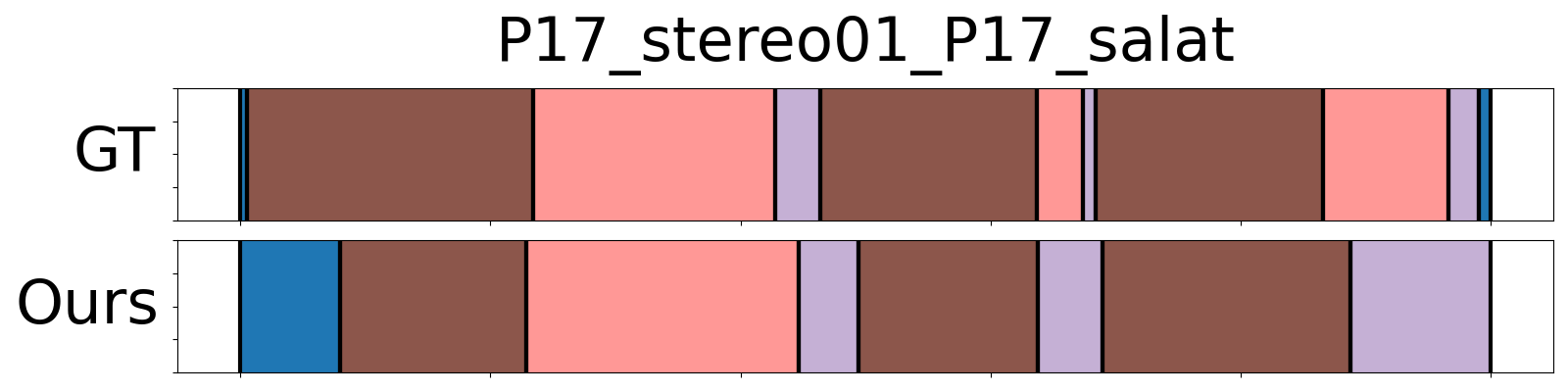}
         \caption{Example videos from the ``salat" activity class within the Breakfast dataset, exhibiting order variations and repeated actions. Both cases are handled by \methodname{}. Actions classes shown are \color{Thistle}\textbf{cut fruit}\color{black}, \color{Brown}\textbf{peel fruit}\color{black}, \color{Purple}\textbf{fruit to bowl}\color{black}, \color{NavyBlue} \textbf{background}.}
         \label{fig:qual_ord_bf}
     \end{subfigure}
     
     \begin{subfigure}[b]{0.35\textwidth}
         \centering
         \includegraphics[width=0.9\textwidth]{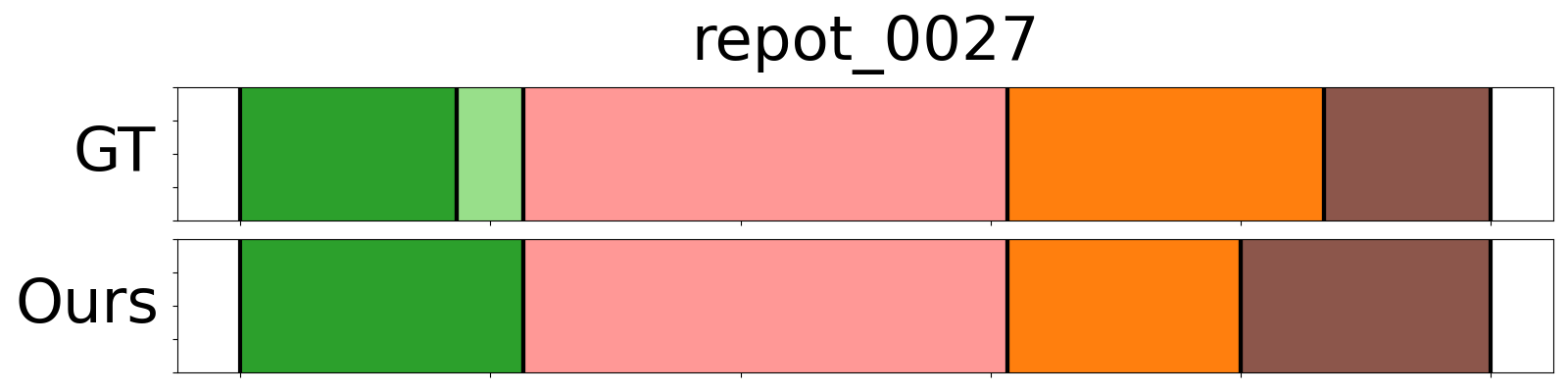}
         \includegraphics[width=0.9\textwidth]{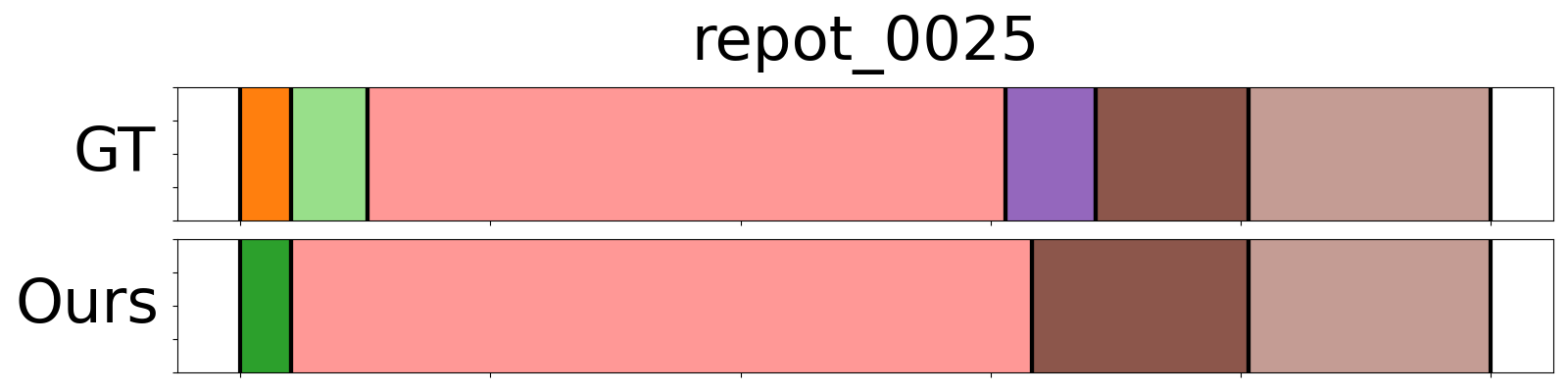}
         \caption{Example videos from YTI (``repot" category). Actions classes shown are \color{ForestGreen}\textbf{tap pot}\color{black}, \color{YellowGreen}\textbf{take plant}\color{black}, \color{CarnationPink}\textbf{loosen root}\color{black}, \color{Orange}\textbf{put soil}\color{black}, \color{Brown}\textbf{add top}\color{black}, \color{Tan}\textbf{water plant}, \color{Plum}\textbf{place plant}\color{black}. }
         \label{fig:qual_ord_yti}
     \end{subfigure}
    \caption{Example action segmentations for videos in the same activity category with differing action orderings. Different colors correspond to different actions. Videos of complex, multi-stage activities can be exhibit markedly different action orderings.}
    \label{fig:qual_order}
\end{figure}
\paragraph{Evaluation Metrics.} 
We follow an evaluation protocol consistent with prior works on the unsupervised segmentation task~\cite{Kukleva_2019_CVPR, Li_2021_CVPR, Kumar_2022_CVPR, VidalMata_2021_WACV, ude}, which we will now describe. Since no ground truth action labels are used during training, we perform Hungarian matching to match learned action clusters to ground truth actions using frame labels and predictions. We evaluate methods under two settings, where the Hungarian matching is performed \textit{per video} (Unsup. per) and across the \textit{full dataset} (Unsup. full). Matching per video does not require clusters to match to the same ground truth action across videos and yields more favorable results.

We comprehensively evaluate all methods using mean-over-frames (MoF), F1-score and mean intersection-over-union (mIoU). MoF measures the percentage of correct \textit{per-frame} predictions and is susceptible to class imbalance issues in a dataset. The F1-score is defined on a \textit{per-segment} basis, where a true positive is defined as having over 50\% of the frames in a ground truth segment predicted correctly. The F1-score is complementary to MoF since large segments (from dominant classes) are treated a single segment. Finally, mIoU computes the IoU averaged over all classes, and is again complementary to both MoF and F1 because because it explicitly handles class imbalance. 

\paragraph{Comparison Methods.} 
We compare our approach against multiple prior works on unsupervised action segmentation~\cite{Kukleva_2019_CVPR, Li_2021_CVPR, Kumar_2022_CVPR, VidalMata_2021_WACV, ude, tran2023permutationaware}. Common to all of these prior approaches is that post-processing involves solving an inference problem in some underlying hidden Markov model (HMM). The states in the HMM correspond to learned action clusters, and observation likelihoods correspond to frame/cluster similarities. A core assumption around the HMM is that the action ordering is known for all videos, which informs the structure of the transition matrix. 

All approaches apart from~\cite{tran2023permutationaware} assume a fixed action ordering for all videos a priori, whereas~\cite{tran2023permutationaware} estimates the action ordering per video using an action transcript prediction module. However, our unsupervised learning pipeline uses \methodname{} for post-processing and does not need these restrictive assumptions or additional learned modules. 

We also add a comparison to unsupervised methods which perform Hungarian matching and evaluate \textit{per video} (Unsup. per)~\cite{Sarfraz_2021_CVPR, Du_2022_CVPR}. For our method, we can use the clusters and frame features learned over the full dataset but evaluate metrics using Hungarian matching per video. TW-FINCH (TWF)~\cite{Sarfraz_2021_CVPR} applies clusters time-weighted frame features within a single video and uses cluster assignments for segmentation. ABD~\cite{Du_2022_CVPR} uses a change detection approach applied over image similarities computed from adjacent video frames. As discussed previously, per video matching tends to yield (much) more favorable results.

\setlength{\tabcolsep}{0.8mm}\begin{table*}[t!]
\centering
\begin{tabular}{llccccc}
\hline
               &                                     & Breakfast                              & YouTube Instr.                         & 50 Salads (Mid)                        & 50 Salads (Eval)                       & Desktop Ass.                           \\ \hline
               &                                     & MoF / F1 / mIoU                        & MoF / F1 / mIoU                        & MoF / F1 / mIoU                        & MoF / F1 / mIoU                        & MoF / F1 / mIoU                        \\ \hline
Per Video   & TWF~\cite{Sarfraz_2021_CVPR}                                 & 62.7 / 49.8 / \textbf{42.3}                     & 56.7 / 48.2 / \,\,\, - \,\,\,                      & \underline{66.8} / \textbf{56.4} / \textbf{48.7}                       & \textbf{71.7} /  \,\,\, - \,\,\, / \,\,\, - \,\,\,                        & \underline{73.3} / \underline{67.7} / \textbf{57.7}                        \\
               & ABD~\cite{Du_2022_CVPR}                                 & \textbf{64.0} / \underline{52.3} / \,\,\, - \,\,\,                      & \underline{67.2} / \underline{49.2} / \,\,\, - \,\,\,                      & \textbf{71.8} / \,\,\, - \,\,\, / \,\,\, - \,\,\,                       & \underline{71.2} /  \,\,\, - \,\,\, / \,\,\, - \,\,\,                       & \,\,\, - \,\,\, / \,\,\, - \,\,\, / \,\,\, - \,\,\,                         \\
               & {\color[HTML]{3531FF} \methodname{} (Ours)}   & \underline{63.3} / \textbf{53.5} / \underline{35.9} & \textbf{71.2} / \textbf{63.3} / 47.8 & 64.3 / \underline{51.1} / \underline{33.4} & 64.5 / 58.9 / 33.0 & \textbf{73.4} / \textbf{68.0} / \underline{47.6} \\ \hline
Full & CTE$^\ast$~\cite{Kukleva_2019_CVPR}                                 & 41.8 / 26.4 / \,\,\, - \,\,\,                      & 39.0 / 28.3 / \,\,\, - \,\,\,                      & 30.2 / \,\,\, - \,\,\, / \,\,\, -  \,\,\,                      & 35.5 / \,\,\, - \,\,\, / \,\,\, - \,\,\,                       & 47.6 / 44.9 / \,\,\, - \,\,\,                      \\
               & CTE$^\dagger$~\cite{Kukleva_2019_CVPR}                                 & 47.2 / 27.0 /  \underline{14.9}                      & 35.9 / 28.0 / \, \underline{9.9} \,                       & 30.1 / 25.5 / \underline{17.9}                      & 35.0 / 35.5 / \underline{21.6}                       & \,\,\, - \,\,\, / \,\,\, - \,\,\, / \,\,\, - \,\,\,                      \\
               & VTE~\cite{VidalMata_2021_WACV}                                 & 48.1 / \,\,\,  -  \,\,\, / \,\,\,  - \,\,\,                    & \,\,\, - \,\,\, / 29.9 / \,\,\,  - \,\,\,                       & 24.2 / \,\,\,  - \,\,\, /\,\,\,  - \,\,\,                      & 30.6 / \,\,\, - \,\,\, / \,\,\, - \,\,\,                       & \,\,\, - \,\,\, / \,\,\, - \,\,\, / \,\,\, - \,\,\,                         \\
               & UDE~\cite{ude}                                 & 47.4 / 31.9 / \,\,\, - \,\,\,                      & 43.8 / 29.6 / \,\,\, - \,\,\,                      & \,\,\, - \,\,\, / \,\,\, - \,\,\, / \,\,\, - \,\,\,                         & 42.2 / 34.4 / \,\,\, - \,\,\,                      & \,\,\, - \,\,\, / \,\,\, - \,\,\, / \,\,\, - \,\,\,                         \\
               & ASAL~\cite{Li_2021_CVPR}                                & \underline{52.5} / 37.9 / \,\,\, - \,\,\,                      & 44.9 / 32.1 / \,\,\, - \,\,\,                      & 34.4 / \,\,\, - \,\,\, / \,\,\, - \,\,\,                       & 39.2 / \,\,\, - \,\,\, / \,\,\, - \,\,\,                       & \,\,\, - \,\,\, / \,\,\, - \,\,\, / \,\,\, - \,\,\,                          \\
               & TOT~\cite{Kumar_2022_CVPR}                                 & 47.5 / 31.0 / \,\,\, - \,\,\,                      & 40.6 / 30.0 / \,\,\, - \,\,\,                        & 31.8 / \,\,\, - \,\,\, / \,\,\, - \,\,\,                        & 47.4 / 42.8 / \,\,\, - \,\,\,                       & 56.3 / 51.7 / \,\,\, - \,\,\,                        \\
               & TOT+~\cite{Kumar_2022_CVPR}                           & 39.0 / 30.3 / \,\,\, - \,\,\,                      & 45.3 / \underline{32.9} / \,\,\, - \,\,\,                      & 34.3 / \,\,\, - \,\,\, /  \,\,\, - \,\,\,                     & 44.5 / 48.2 / \,\,\, - \,\,\,                       & 58.1 / 53.4 / \,\,\, - \,\,\,                      \\
               & UFSA (M)~\cite{tran2023permutationaware}                           & \,\,\, - \,\,\, / \,\,\, - \,\,\, / \,\,\, - \,\,\,                      & 43.2 / 30.5 / \,\,\, - \,\,\,                      & \,\,\, - \,\,\, / \,\,\, - \,\,\, / \,\,\, - \,\,\,                    & 47.8 / 34.8 / \,\,\, - \,\,\,                       & \,\,\, - \,\,\, / \,\,\, - \,\,\, / \,\,\, - \,\,\,                      \\
               & UFSA (T)~\cite{tran2023permutationaware}                           & 52.1 / \underline{38.0} / \,\,\, - \,\,\,                      & \underline{49.6} / 32.4 / \,\,\, - \,\,\,                      & \underline{36.7} / \underline{30.4} /  \,\,\, - \,\,\,                     & \underline{55.8} / \underline{50.3} / \,\,\, - \,\,\,                       & \underline{65.4} / \underline{63.0} / \,\,\, - \,\,\,                      \\
               & {\color[HTML]{3531FF} \methodname{} (Ours)} & \textbf{56.1} / \textbf{38.3} / \textbf{18.6}                     & \textbf{52.9} / \textbf{35.1} / \textbf{24.7}                     & \textbf{46.2} / \textbf{37.4} / \textbf{24.9}                     & \textbf{59.3} / \textbf{53.6} / \textbf{30.1}                     & \textbf{70.4} / \textbf{68.0} / 45.9                     \\ \hline
\end{tabular}
\caption{Summary of experimental results. For all evaluation metrics, higher is better and \textbf{bold} (respectively,~\underline{underline}) indicates the best (respectively second best) performing methods. A ``-" indicates that a metric was not reported in the original paper. Note that ``Full" and ``Per Video" relate to the evaluation metrics being computed by applying Hungarian matching on a whole dataset and per video basis, respectively. For CTE~\cite{Kukleva_2019_CVPR}, $\ast$ and $\dagger$ indicate results reported and reproduced using model checkpoints provided by the authors, respectively. Results were generated using one run, consistent with prior works, however we provide results for 5 runs for FS and DA in the appendix.}\label{tab:main}
\end{table*}

\begin{table*}[t!]
\centering
\begin{tabular}{lccccc}
\hline
                      & Breakfast          & YouTube Instr.     & 50 Salads (Mid)    & 50 Salads (Eval)   & Desktop Ass.       \\ \hline
                      & MoF / F1 / mIoU    & MoF / F1 / mIoU    & MoF / F1 / mIoU    & MoF / F1 / mIoU    & MoF / F1 / mIoU    \\ \hline
Base                  & 56.1 / 38.3 / 18.6 & 52.9 / 35.1 / 24.7 & 46.2 / 37.4 / 24.9 & 59.3 / 53.6 / 30.1 & 70.4 / 68.0 / 45.9 \\
No $k$-means          & 57.7 / 36.0 / 17.1 & 49.5 / 34.1 / 21.3 & 42.1 / 34.5 / 22.1 & 55.2 / 52.7 / 28.2 & 56.8 / 61.7 / 35.6 \\
No temp. prior        & 48.5 / 24.9 / 16.3 & 46.9 / 26.2 / 14.6 & 43.1 / 36.5 / 21.9 & 59.7 / 54.7 / 26.3 & 31.4 / 18.7 / 10.6 \\
Balanced OT           & 29.7 / 29.3 / 17.8 & 39.4 / 31.4 / 14.6 & 39.7 / 39.8 / 25.3 & 35.7 / 41.8 / 24.9 & 56.5 / 72.7 / 37.8 \\
No GW (train)         & 34.4 / 25.9 / 14.4 & 41.1 / 24.9 / 11.7 & 29.0 / 22.6 / 14.3 & 35.1 / 38.6 / 22.5 & 49.5 / 49.4 / 30.4 \\
No GW (train \& test) & 32.6 / 21.3 / 10.4 & 38.0 / 21.8 / 11.9 & 17.3 / \,\,\,4.2 / \,\,\,8.9 & 25.8 / 21.3 / 14.8 & 45.4 / 42.0 / 26.5 \\ \hline
\end{tabular}
\caption{Ablation study results for the Unsup (full) setting. Effects are not additive. For all evaluation metrics, higher is better.}
\label{tab:ablation}
\end{table*}

\section{Results}\label{sec:results}

\subsection{State-of-the-Art Comparison}

Our experimental results are presented in Tab.~\ref{tab:main}. For the Unsup. (full) setting, our method consistently outperforms all relevant prior works across all evaluation metrics. For the Breakfast dataset, our method yields comparable results to ASAL~\cite{Li_2021_CVPR} and the recent joint representation learning and clustering approach UFSA~\cite{tran2023permutationaware}, where \textit{UFSA (T)} in Tab.~\ref{tab:main} refers to the authors' full model with a transformer frame encoder. However, we show major improvements over \textit{UFSA (T)} for the FS (Mid) and DA datasets. 

We note that our method uses a simple MLP frame encoder, whereas UFSA uses a more complex transformer encoder, on top of multiple additional learned modules. For a fairer comparison, we also present results for UFSA with a comparable MLP encoder, displayed as \textit{UFSA (M)} in Tab.~\ref{tab:main}. Compared to UFSA (M), we show significant improvements for YTI and FS (Eval). Similar to UFSA, we believe our method's ability to handle out-of-order alignments during segmentation contributes to the improved performance. Fig.~\ref{fig:qual_order} contains qualitative examples where actions are seen in different orders across videos within a dataset.

For the per video evaluation protocol, our method outperforms TWF~\cite{Sarfraz_2021_CVPR} and ABD~\cite{Du_2022_CVPR} on the BF and YTI datasts for MoF and F1, however for BF it performs poorly on mIoU. We believe this is because our method is more successfully finding segmentation boundaries for large classes, at the cost of failing to identify shorter, more underrepresented actions. In addition, our method appears to underperform on FS (Mid and Eval) and yield comparable results for DA. We emphasize that the results were optimized for the Unsup. (full) setting, where actions should be identified correctly across videos. An interesting avenue for future work is to investigate how to estimate action clusters using frame features from a \textit{single video} for our method.

\begin{figure}[t!]
     \centering
     \begin{subfigure}[b]{0.20\textwidth}
         \centering
         \includegraphics[width=\textwidth]{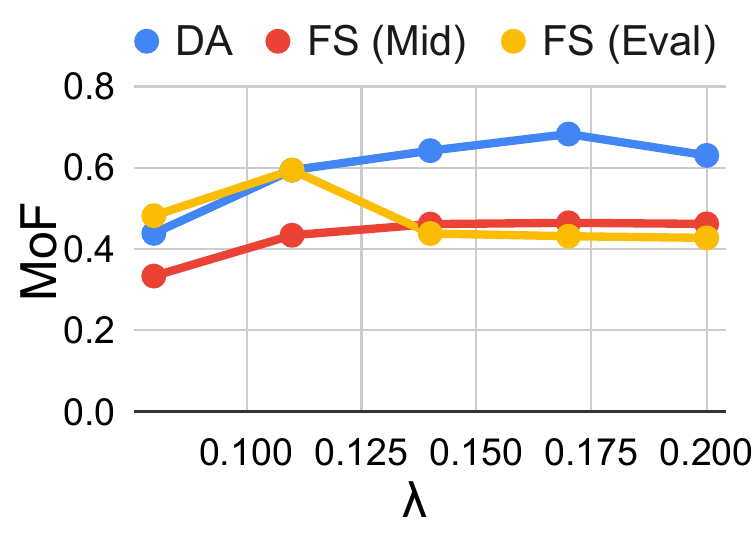}
         \caption{Balanc. assign. weight $\lambda$}
         \label{fig:sens_lambda}
     \end{subfigure}
     \begin{subfigure}[b]{0.20\textwidth}
         \centering
         \includegraphics[width=\textwidth]{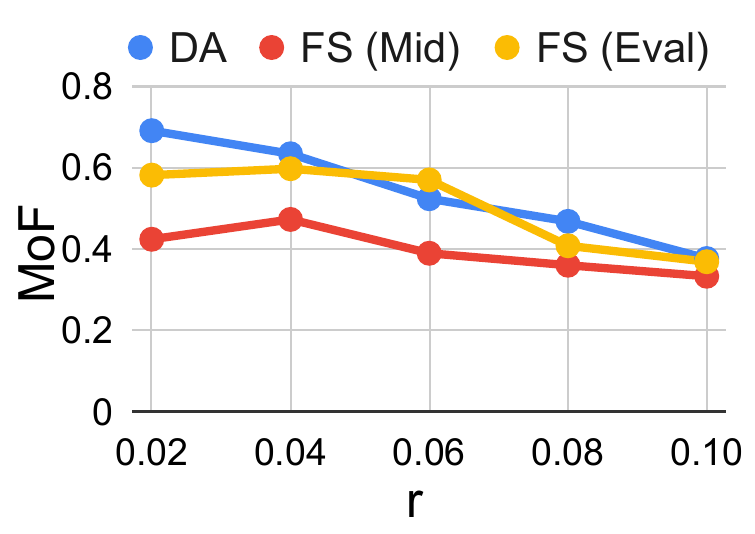}
         \caption{GW radius $r$}
         \label{fig:sens_r}
     \end{subfigure}
     
     \begin{subfigure}[b]{0.20\textwidth}
         \centering
         \includegraphics[width=\textwidth]{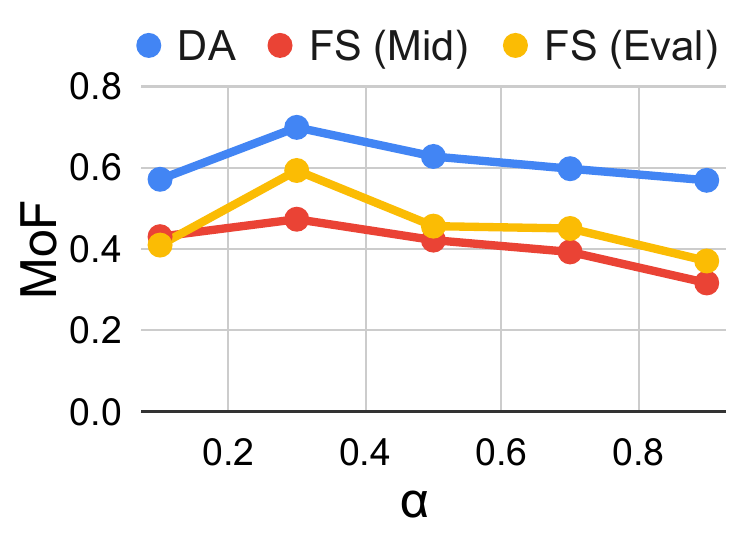}
         \caption{GW weight $\alpha$}
         \label{fig:sens_alpha}
     \end{subfigure}
     \begin{subfigure}[b]{0.20\textwidth}
         \centering
         \includegraphics[width=\textwidth]{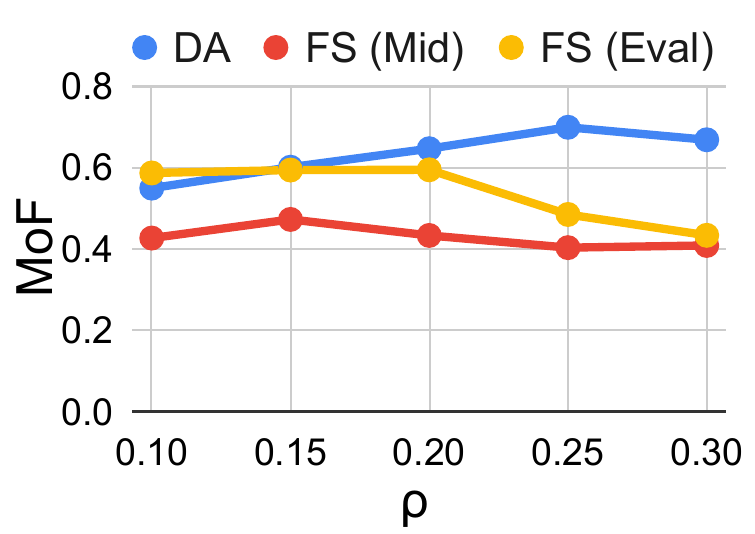}
         \caption{Global temporal prior $\rho$}
         \label{fig:sens_rho}
     \end{subfigure}

     \begin{subfigure}[b]{0.20\textwidth}
         \centering
         \includegraphics[width=\textwidth]{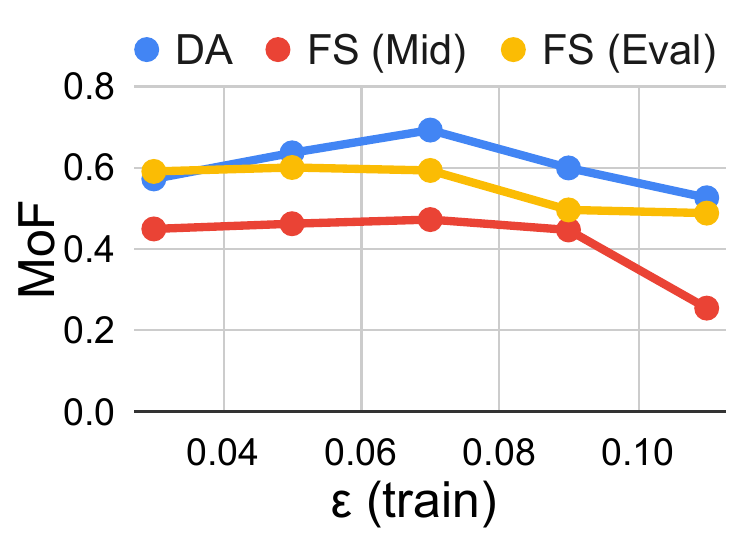}
         \caption{Entropy reg. $\epsilon_{\text{train}}$}
         \label{fig:sens_eps}
     \end{subfigure}
     \begin{subfigure}[b]{0.20\textwidth}
         \centering
         \includegraphics[width=\textwidth]{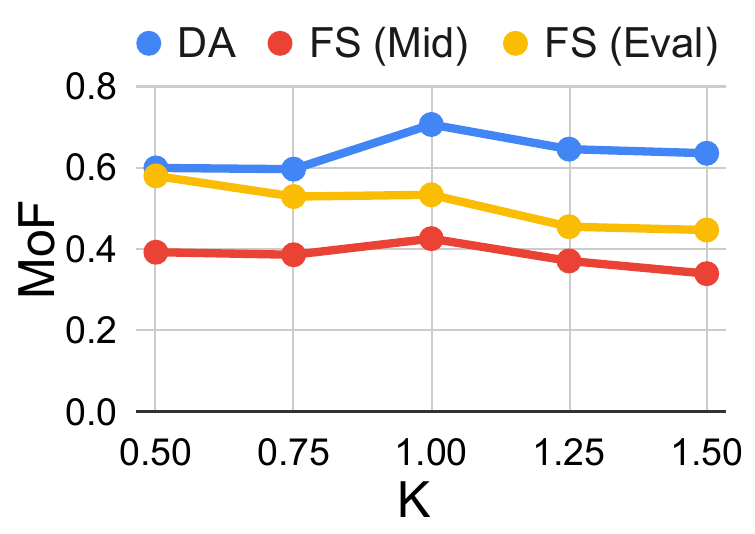}
         \caption{Ratio of learned/true actions}
         \label{fig:sens_K}
     \end{subfigure}
    \caption{Sensitivity analysis on FUGW OT hyperparameters.}
    \label{fig:sensitivity}
\end{figure}

\subsection{Ablation Study}\label{sec:ablation}

In this section, we analyze the effects of various choices in our learning pipeline and present the results in Table~\ref{tab:ablation}. 

\paragraph{Temporal prior and $k$-means.} 
First, we ablate the effect of initialization of action clusters by switching $k$-means to random sampling (\textit{No $k$-means}). We find that overall, using $k$-means is beneficial because the clusters are initialized to be more evenly represented within videos, yielding higher quality pseudo-labels at the start of training. Second, we remove the temporal prior (\textit{No temp. prior}) described in Sec.~\ref{sec:replearning}. The prior appears to yield a positive effect overall.

\begin{figure}[t!]
     \centering
     \begin{subfigure}[b]{0.33\textwidth}
         \centering
         \includegraphics[width=0.8\textwidth]{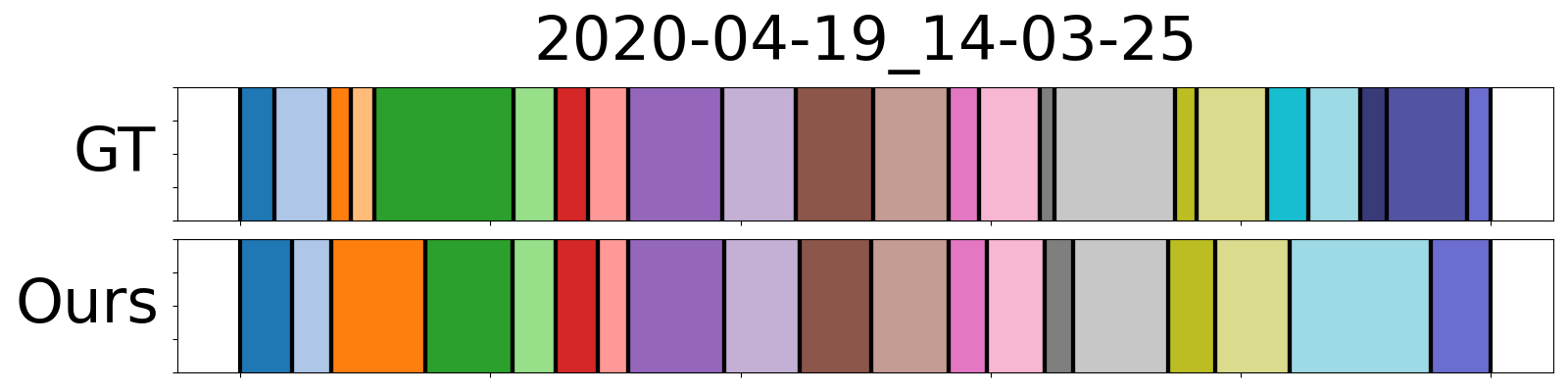}
         \caption{Desktop Assembly}
         \label{fig:qual_da_main}
     \end{subfigure}
     
     \begin{subfigure}[b]{0.33\textwidth}
         \centering
         \includegraphics[width=0.8\textwidth]{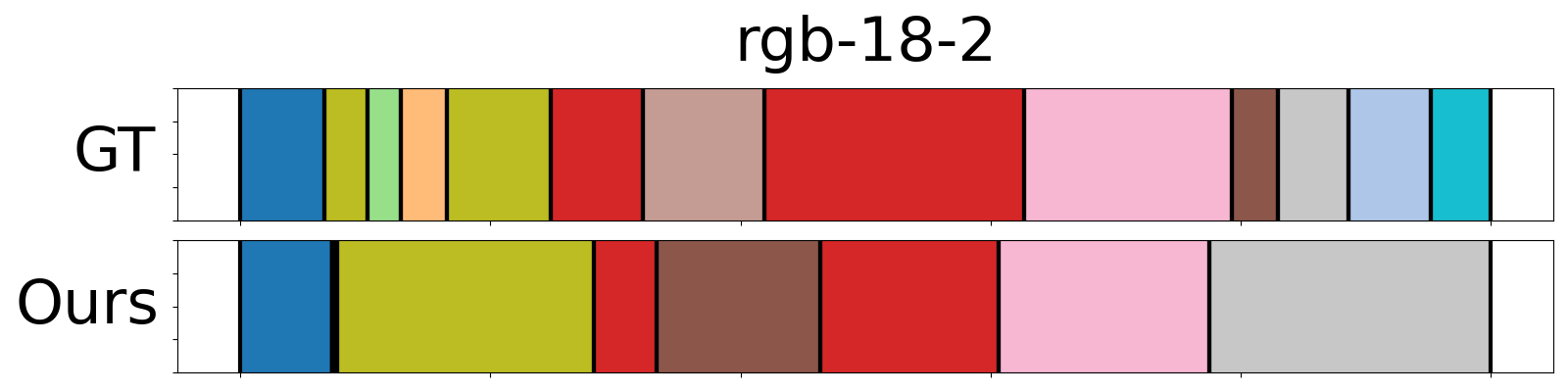}
         \caption{50 Salads (Eval)}
         \label{fig:qual_fs_main}
     \end{subfigure}
    \caption{Example segmentations from DA and FS (Eval). Note the dominant red class in FS, whereas DA is (roughly) balanced.}
    \label{fig:qual_sens}
\end{figure}

\paragraph{Effect of (un)balanced assignment.} 
For \methodname{}, we first analyze the effect of forcing a balanced assignment to actions (Balanced OT), ignoring the unbalanced OT formulation described in Sec.~\ref{ssec:unbalancedvideo}. For the BF, YTI and FS datasets, we observed that dominant classes are common in the ground truth annotations, and forcing a balanced assignment to actions severely degrades performance overall. However for DA, enforcing a balanced assignment actually slightly improves performance. This is because the ground truth actions are (relatively) evenly represented across the DA dataset. Compare Figures~\ref{fig:qual_fs_main} and~\ref{fig:qual_order} to Figure~\ref{fig:qual_da_main} to see class balance discrepancies across datasets. 

\paragraph{Effect of GW structural prior.} 
Finally, we remove the GW problem described in Sec.~\ref{ssec:gwot} at training time only (\textit{No GW (train)}), as well as at both training and test time (\textit{No GW (train \& test)}). We observe that removing the temporal consistency of the pseudo-labels alone is enough to compromise the performance of our method. As expected, also removing the temporal consistency of the test time segmentations further reduces performance.

\subsection{Sensitivity Analysis}

In addition to the ablation studies, we performed a sensitivity analysis on the hyperparameters of our action segmentation OT method during training. These hyperparameters strongly influence the behavior of the pseudo-labels and resultant learned representations. We illustrate the results in Fig.~\ref{fig:sensitivity} and perform this analysis using the desktop assembly (DA) and 50 Salads (FS) datasets at both the Eval and Mid granularity, using MoF as the metric. 

\paragraph{Effect of $\lambda$, $r$ and $\alpha$.} 
As discussed in Sec.~\ref{sec:ablation}, a (more) balanced assignment benefits datasets such as DA and FS (Mid) with a large number of action classes. Unsurprisingly, MoF improves for DA and FS (Mid) for increasing $\lambda$, whereas FS (Eval), which has more dominant action classes, has an optimal point around $\lambda=0.11$. Furthermore, the GW radius $r$ is positively correlated to the resulting segment length. We can see that a small value of $r=0.02$ is enough for temporal consistency in FS (Mid and Eval) and DA, yielding the best MoF. Finally, MoF is relatively insensitive to the GW structure weight $\alpha$, except at the extremes. We find a value of $\alpha\in[0.2, 0.5]$ to work well across all evaluated datasets.

\paragraph{Effect of $\rho$, $\epsilon_\text{train}$ and $K$.} 
Finally, we observe that higher values for the global temporal prior weight $\rho$ benefits DA. We believe this is because a (banded) diagonal coupling is a reasonable solution for datasets such as DA with relatively balanced action classes and identical action orderings across videos. However, FS (Eval and Mid) seem to benefit from removing the global prior altogether. The pseudo-labels at initialization appear to be high quality, and do not need additional guidance from the global structure. 

Furthermore, we observe that the optimal value for entropy regularization for the OT problem, denoted $\epsilon_\text{train}$, should be set between 0.06 and 0.08. Low values for $\epsilon_\text{train}$ may result in numerical instability arising from the solver, while high values may result in incoherent segmentations. Finally, for datasets large number of classes (DA and FS Mid), setting number the of clusters $K$ to be equal to true number of actions $K_\text{gt}$ appears to be optimal. For FS (Eval) where dominant classes as present, performance actually improves by setting fewer clusters. 

\subsection{Post-Processing for Supervised Methods}

We show the generality of \methodname{} as a post-processing method by applying it to the outputs of the supervised method MS-TCN++~\cite{mstcnplus} for the 50 Salads dataset. We use the recommended multi-scale configuration (MS) with three refinement layers as well as without refinement layers, i.e., single-scale (SS). \cite{mstcnplus} observed that the primary function of the refinement layers is to improve the temporal consistency of the predictions. Our results verify this finding, since SS with \methodname{} yields comparable performance with MS. Furthermore, applying \methodname{} to MS yields significant further improvements. See the appendix for hyperparameter settings and more details on this experiment.

\begin{table}[h]
\setlength{\tabcolsep}{0.8mm}\begin{tabular}{  c  c  c  c  c  c  c  }
 \hline
  & Accuracy & ED & F1@10 & F1@25 & F1@50 \\ \hline
MS & 83.5 & 72.6  & 80.1  & 77.2 & 69.4   \\ 
SS & 76.0 & 36.3  & 44.8 & 42.3 & 34.9 \\ 
 \hline
MS + \methodname{} & 83.6 & 77.6 & 84.4 & 83.1 & 74.9 \\
SS + \methodname{} & 78.3 & 72.5 & 80.3 & 78.5 & 69.9 \\ \hline
\end{tabular}
\caption{ED and F1@$N$ refers to the segmental edit distance and F1 score at overlapping threshold of $N\%$, respectively. Acc. is the MoF accuracy. Higher is better for all metrics.}
\label{fig:sup_asot}
\end{table}

\section{Discussion and Conclusion}

In this paper we present a novel method for the unsupervised action segmentation task on long, untrimmed videos. Our post-processing approach \methodname{}, yields temporally consistent segmentations without prior knowledge of the action ordering, required by previous approaches. Furthermore, we show that in the unsupervised setting, \methodname{} produces pseudo-labels suitable for self-training. Future work will investigate the semi and fully-supervised action segmentation settings, which will require backpropagating through our \methodname{} problem. Finally, developing methods for unsupervised learning over all activity categories for challenging datasets such as Breakfast is an important direction for future work, and will require a more sophisticated learning pipeline than the one proposed in Sec.~\ref{sec:replearning}.
\paragraph{Acknowledgements.} This work was supported by an Australian Research Council (ARC) Linkage grant (project number LP210200931). Ming Xu thanks Akshay Asthana and Liang Zheng for helpful discussions about this work.

{\small
\bibliographystyle{ieee_fullname}
\bibliography{egbib}
}

\clearpage \appendix 
\section{Pseudo-code for \methodname{}}\label{sec:pseduocode}

In Alg.~\ref{alg:pseudocode}, we present pseudo-code around the numerical solver for our action segmentation optimal transport (\methodname{}) algorithm. Our approach is based on projected mirror descent, similar to Peyre \etal~\cite{icml_16_peyre_gw}. We initialize the coupling $\mathbf{T}$ element-wise as $\mathbf{T}_{ij} = 1 / (NK)$. Each iteration then involves two steps, given by

\begin{enumerate}
    \item \textbf{Update step (Lines~6--16):} The first-order update step is computed as a mirror descent step under the KL-divergence. The update step at iteration $t$ is given by $\hat{\mathbf{T}}^{t+1} = \mathbf{T}^t \odot \exp(-\phi \nabla_\mathbf{T}^t\mathcal{F}_\text{ASOT}(\mathbf{C}, \mathbf{T}^t))$, where $\phi > 0$ is a step size and
    \begin{align}
    \mathcal{F}_\text{ASOT}(\mathbf{C}, \mathbf{T}, \epsilon) \coloneqq &\mathcal{F}_\text{FGW}(\mathbf{C}, \mathbf{T}) + \nonumber \\
    &\lambda \text{D}_\text{KL} (\mathbf{T}^\top \mathbf{1}_n \| \mathbf{q}) 
     - \epsilon H(\mathbf{T}).
    \end{align}
    \item \textbf{Projection step (Lines~10--15):} The projection step involves ensuring the polytope constraints are satisfied. This involves simply rescaling the rows of $\mathbf{T}$, to satisfy the frame-wise marginal constraint, \ie $T^{t+1}_{ij} = p_i \hat{T}_{ij}^{t+1} / (\hat{\mathbf{T}}^{t+1}\mathbf{1}_K)_i$.
\end{enumerate}



\begin{algorithm}
    \caption{Action Segmentation Optimal Transport}
    \label{alg:pseudocode}
    \begin{algorithmic}[1]
        \Require Video frame to action class affinity matrix $\mathbf{C}^k\in\mathbb{R}^{N\times K}$, derived from a video encoder and hyperparameters $\alpha\in[0,1]$, $r \in[0,1]$, $\lambda \geq 0$, $\epsilon > 0$, $\phi > 0$, $n_\text{iter} > 0$.
        \Ensure Soft assignment probabilities $\mathbf{T}^\star \in \mathbb{R}^{N \times K}_{+}$.
        \Procedure{\methodname{}}{$\mathbf{C}^k$, $\alpha$, $r$, $\lambda$, $\epsilon$, $\phi$, $n_\text{iter}$}
            \State $\mathbf{p} \gets \frac{1}{N} \mathbf{1}_N$ \Comment{Frame marginals}
            \State $\mathbf{q} \gets \frac{1}{K} \mathbf{1}_K$ \Comment{Balanced action marginals}
            \State Construct $\mathbf{C}^v$, $\mathbf{C}^a$ using \eqref{eq:cv_def}.
            \State $\mathbf{C} = \{\mathbf{C}^k, \mathbf{C}^v, \mathbf{C}^a\}$
            \State $\mathbf{T}^{0} \gets \mathbf{p} \otimes \mathbf{q}$ \Comment{$\otimes$ denotes the outer product}
            \For{$t \gets 0$ to $n_\text{iter}-1$} 
                \State $\hat{\mathbf{T}}^{t+1} = \mathbf{T}^t \odot \exp(-\phi \nabla_{\mathbf{T}}\mathcal{F}_\text{ASOT}(\mathbf{C}, \mathbf{T}^t))$
                \State $T^{t+1}_{ij} = p_i \hat{T}_{ij}^{t+1} / (\hat{\mathbf{T}}^{t+1}\mathbf{1}_K)_i \quad \forall i, j$
            \EndFor
            \State \textbf{return} $\mathbf{T}^{n_\text{iter}}$
        \EndProcedure
    \end{algorithmic}
\end{algorithm}

We provide a full reference implementation for \methodname{} and training code at \url{https://github.com/mingu6/action_seg_ot}.

\section{Additional Details for Unsupervised Experiment}

For both the training (pseudo-labelling) and inference phases, we use 25 iterations for projected mirror descent (see Sec.~\ref{sec:pseduocode}). We use the same Gromov-Wasserstein (GW) radius parameter $r$ across training and inference, which is set at 0.04 for all datasets except for desktop assembly, which is set at 0.02. We observed that desktop assembly has smaller ground truth segments relative to video length, benefiting a lower value for $r$.

\paragraph{ASOT pseudo-labelling.} The \methodname{} hyperparameters used to generate pseudo-labels during the training phase of our unsupervised action segmentation pipeline are described as follows. The GW structure weight $\alpha$ is set to 0.3 for all datasets except for Breakfast where $\alpha=0.4$ and furthermore, entropy regularization weight $\epsilon=0.07$ for all datasets. Table~\ref{tab:hyperparams} presents the remaining hyperparameters. 

\paragraph{ASOT inference.} The hyperparameters used to generate segmentations from embeddings learned from our unsupervised learning pipeline are described as follows. First, we set the unbalanced weight $\lambda$ to a low value (0.01) for all datasets. At inference, \methodname{} will segment according to the underlying learned embeddings, and will not encourage a more balanced assignment to action classes. Second, we set $\alpha=0.6$ for all datasets except for Breakfast where $\alpha=0.7$, because a higher value for $\alpha$ results in stronger temporal consistency due to heavier weighting of the GW structure objective. Finally, we set $\epsilon=0.04$, since lower levels of entropy regularization yields sharper segmentations. 

\paragraph{Representation learning.} We set the learning rate at $10^{-3}$ and weight decay to $10^{-4}$ for all datasets. We sample a batch of 2 videos, where for each video, we sample 256 frames randomly by partitioning each video into 256 uniform intervals and sampling a single frame from each interval. Our MLP architecture has a single hidden layer with ReLU activations. The hidden layer size is 128 for all datasets, with an output feature dimension of 40, with the exception of YouTube Instructions (YTI), which have sizes 32 and 32, respectively. Output features are $l_2$-normalized before applying \methodname{}. The smaller model size from YTI arises from the significantly higher dimensional input features compared to the remaining datasets.

\begin{table}[ht!]
\centering
\begin{tabular}{lccccc}
\hline
                               &  BF & YTI & FS (M) & FS (E) & DA \\ \hline
    Unbalanced ($\lambda$)      & 0.1      & 0.12           & 0.15            & 0.11             & 0.16         \\
                    Global prior ($\rho$) & 0.2      & 0.15            & 0.15             & 0.15              & 0.25          \\ 
                    Num. epochs & 15 & 10 & 30 & 30 & 30 \\ \hline
\end{tabular}
\caption{Hyperparameter settings for ASOT used to generate pseudo-labels in the \textit{training phase} of our unsupervised action segmentation pipeline. FS (M) and FS (E) represent the 50 Salads Mid and Eval splits, respectively.}
\label{tab:hyperparams}
\end{table}

\section{Experimental Setup (Supervised)}

For the supervised example, the MS-TCN~\cite{mstcnplus} architecture outputs logits $\mathbf{L}\in\mathbb{R}^{N\times K}$ directly instead of video frame and action embeddings. We can transform these logits into a cost matrix $\mathbf{C}^k$ using
\begin{equation}
    C^k_{ij} = 2 \left( 1 - \frac{L_{ij} - L_\text{min}}{L_\text{max} - L_\text{min}}\right),
\end{equation}
where $L_\text{max} \coloneqq \max_{i, j} L_{ij}$ and $L_\text{min} \coloneqq \min_{i, j} L_{ij}$. This is a simple method for converting the logits which model a frame/action affinity into a cost matrix with elements scaled between $[0, 2]$.

We then apply \methodname{} to the cost matrix derived per video with hyperparameters $\lambda = 0.05$, $\alpha = 0.4$, $\epsilon = 0.06$ and $r = 0.01$. We evaluate our method using 5-fold cross validation, similar to~\cite{mstcnplus}.

\section{Additional Sensitivity Analysis}

\begin{figure}[t!]
     \centering
     \begin{subfigure}[b]{0.22\textwidth}
         \centering
         \includegraphics[width=\textwidth]{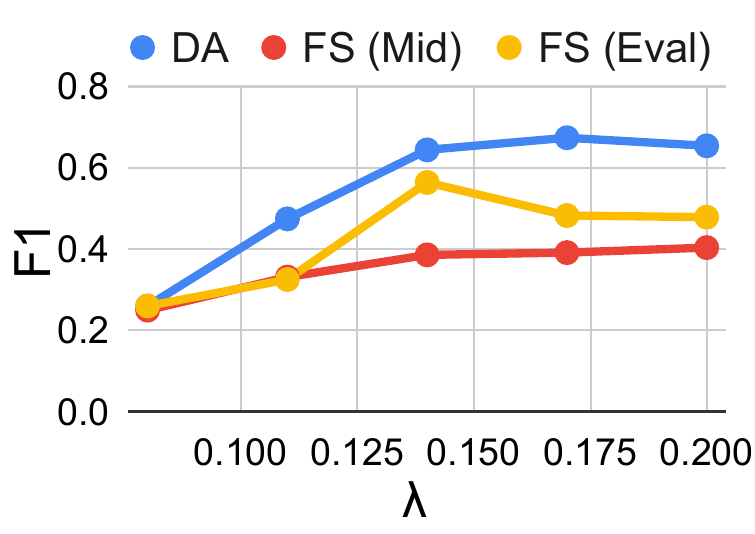}
         \caption{Balanc. assign. weight $\lambda$}
         \label{fig:sens_lambda_f1}
     \end{subfigure}
     \begin{subfigure}[b]{0.22\textwidth}
         \centering
         \includegraphics[width=\textwidth]{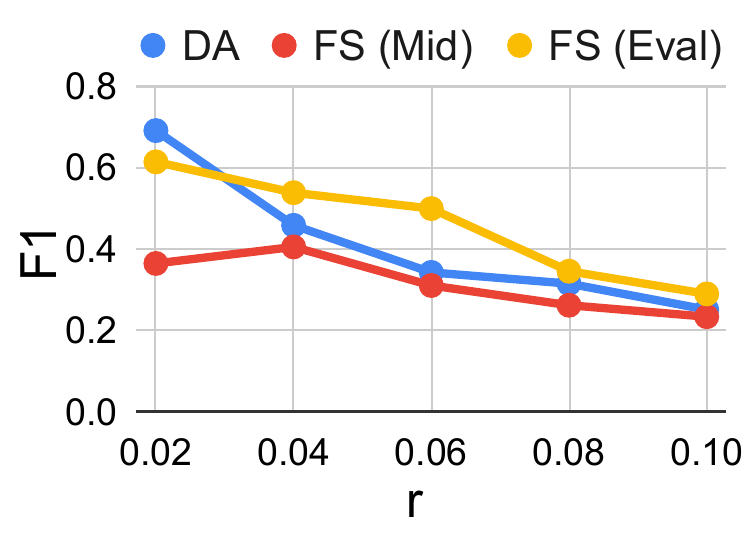}
         \caption{GW radius $r$}
         \label{fig:sens_r_f1}
     \end{subfigure}
     
     \begin{subfigure}[b]{0.22\textwidth}
         \centering
         \includegraphics[width=\textwidth]{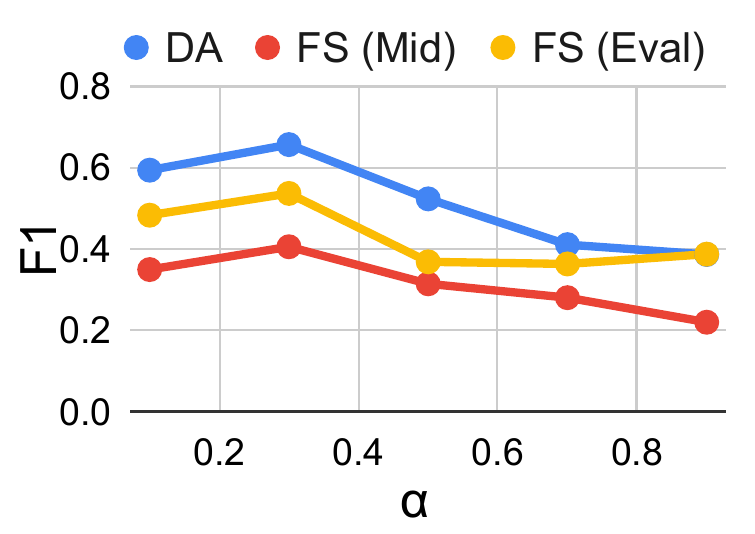}
         \caption{GW weight $\alpha$}
         \label{fig:sens_alpha_f1}
     \end{subfigure}
     \begin{subfigure}[b]{0.22\textwidth}
         \centering
         \includegraphics[width=\textwidth]{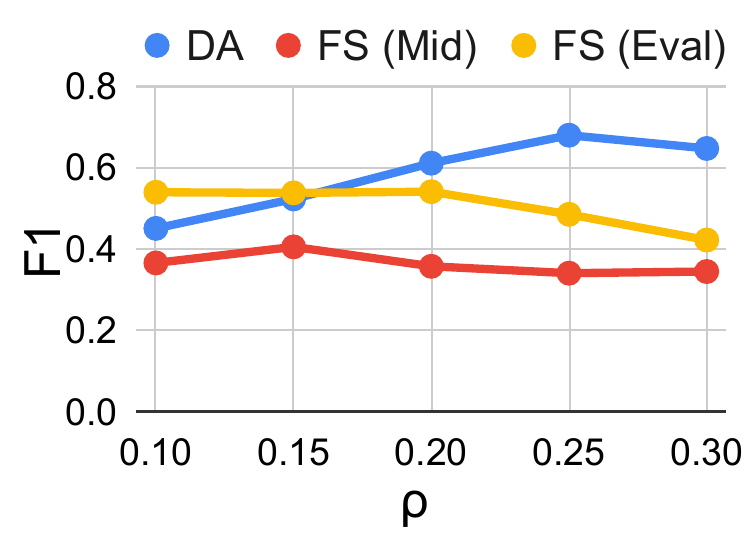}
         \caption{Global temporal prior $\rho$}
         \label{fig:sens_rho_f1}
     \end{subfigure}

     \begin{subfigure}[b]{0.22\textwidth}
         \centering
         \includegraphics[width=\textwidth]{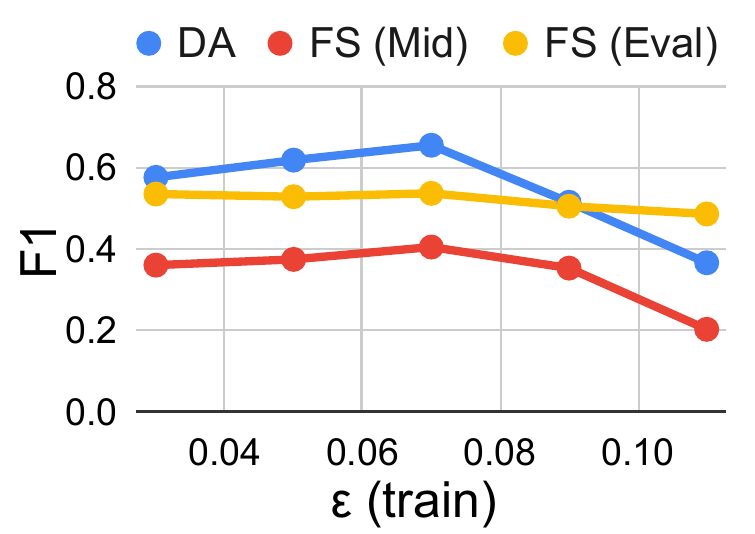}
         \caption{Entropy reg. $\epsilon_{\text{train}}$}
         \label{fig:sens_eps_f1}
     \end{subfigure}
     \begin{subfigure}[b]{0.22\textwidth}
         \centering
         \includegraphics[width=\textwidth]{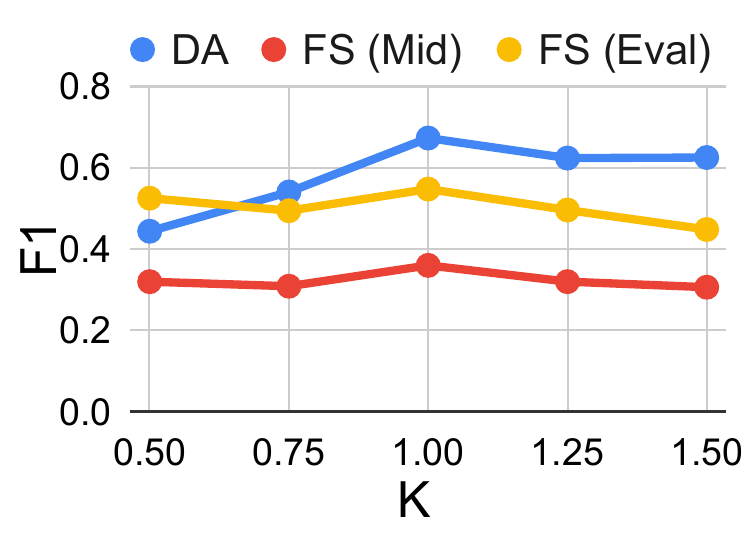}
         \caption{Ratio of learned/true actions}
         \label{fig:sens_K_f1}
     \end{subfigure}
    \caption{Sensitivity analysis reporting F1-score on \methodname{} hyperparameters.}
    \label{fig:sensitivity_f1}
\end{figure}

\begin{figure}[t!]
     \centering
     \begin{subfigure}[b]{0.22\textwidth}
         \centering
         \includegraphics[width=\textwidth]{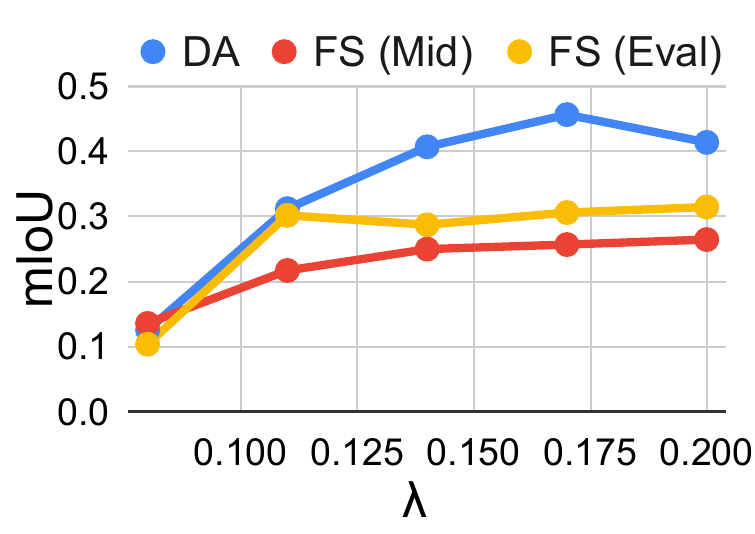}
         \caption{Balanc. assign. weight $\lambda$}
         \label{fig:sens_lambda_miou}
     \end{subfigure}
     \begin{subfigure}[b]{0.22\textwidth}
         \centering
         \includegraphics[width=\textwidth]{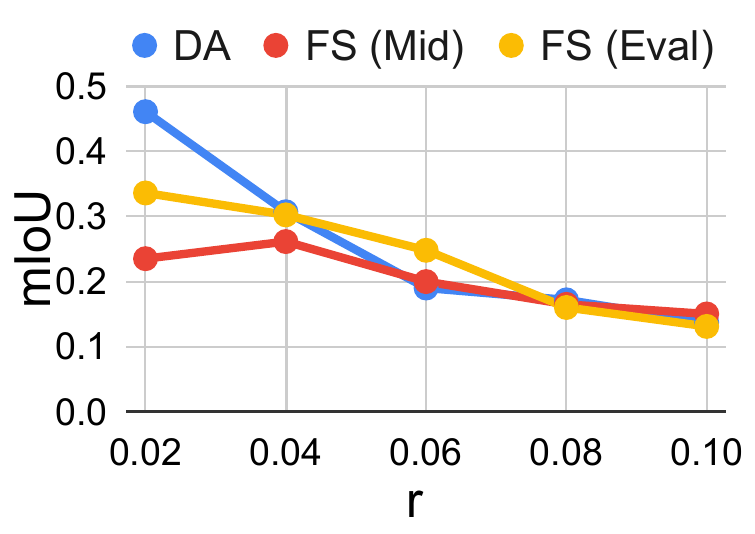}
         \caption{GW radius $r$}
         \label{fig:sens_r_miou}
     \end{subfigure}
     
     \begin{subfigure}[b]{0.22\textwidth}
         \centering
         \includegraphics[width=\textwidth]{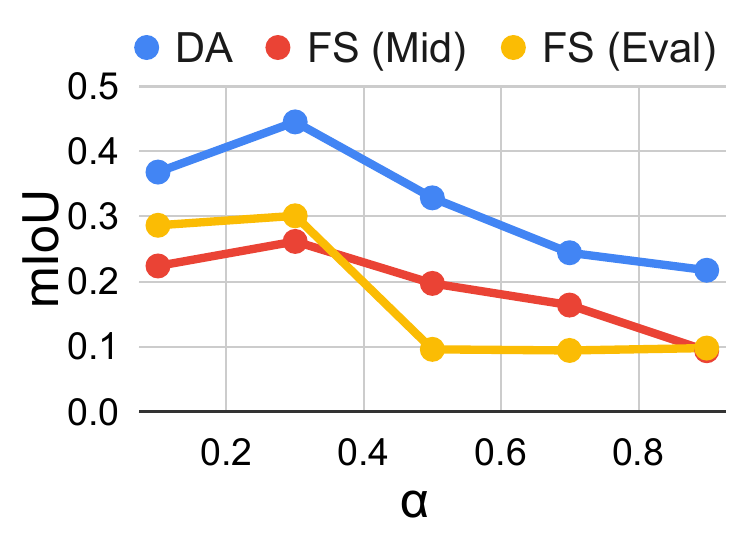}
         \caption{GW weight $\alpha$}
         \label{fig:sens_alpha_miou}
     \end{subfigure}
     \begin{subfigure}[b]{0.22\textwidth}
         \centering
         \includegraphics[width=\textwidth]{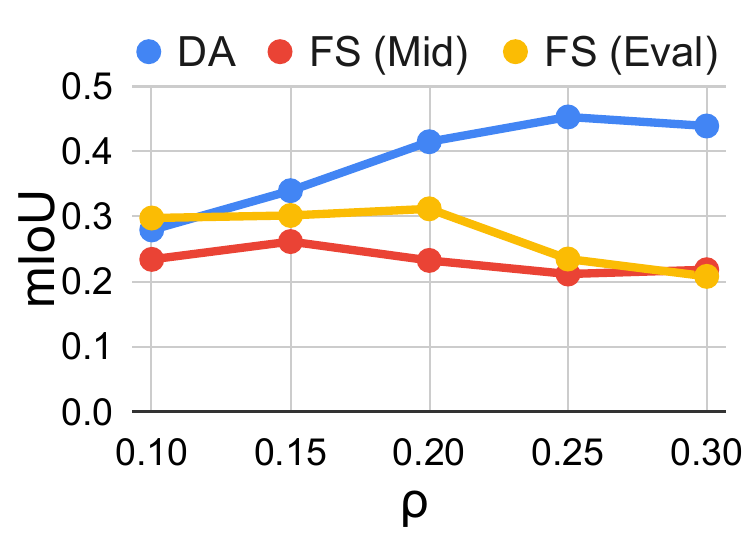}
         \caption{Global temporal prior $\rho$}
         \label{fig:sens_rho_miou}
     \end{subfigure}

     \begin{subfigure}[b]{0.22\textwidth}
         \centering
         \includegraphics[width=\textwidth]{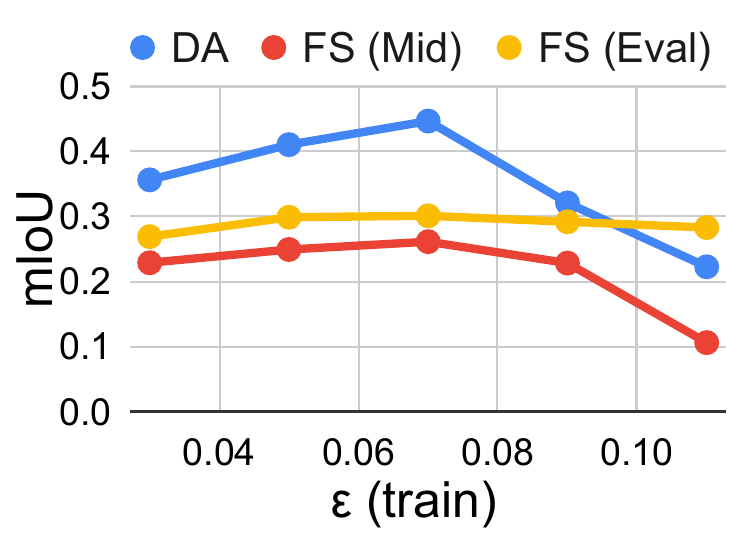}
         \caption{Entropy reg. $\epsilon_{\text{train}}$}
         \label{fig:sens_eps_miou}
     \end{subfigure}
     \begin{subfigure}[b]{0.22\textwidth}
         \centering
         \includegraphics[width=\textwidth]{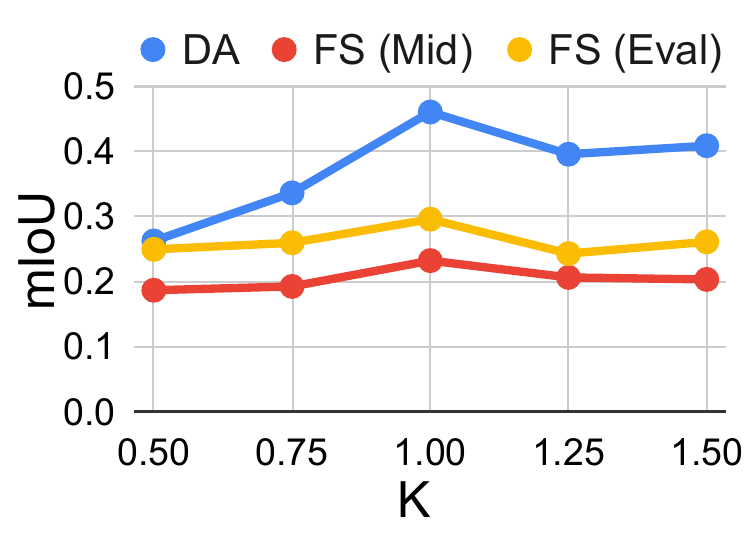}
         \caption{Ratio of learned/true actions}
         \label{fig:sens_K_miou}
     \end{subfigure}
    \caption{Sensitivity analysis reporting mIoU-score on \methodname{} hyperparameters.}
    \label{fig:sensitivity_miou}
\end{figure}

In addition to the sensitivity analysis under the MoF metric presented in Fig.~\ref{fig:sensitivity}, we present additional results for F1-score and mIoU in Fig.~\ref{fig:sensitivity_f1} and~\ref{fig:sensitivity_miou}, respectively. The results follow very similar trends to MoF, however it is notable in Fig.~\ref{fig:sens_lambda_miou} that mIoU performance collapses to almost 0 for $\lambda=0$, whereas MoF does not (see Fig.~\ref{fig:sens_lambda}). This is especially notable for the FS (Eval) dataset. This discrepancy between MoF and mIoU suggests the presence of unbalanced classes within the datasets, especially FS (Eval).

\section{Reproducibility} Our results in Tab.~\ref{tab:main} and~\ref{tab:ablation} were produced using one run, consistent with prior works in unsupervised action segmentation~\cite{Kukleva_2019_CVPR, Li_2021_CVPR, Kumar_2022_CVPR, VidalMata_2021_WACV, ude, tran2023permutationaware}. To show the robustness of our methods, we used five runs on the 50 Salads and Desktop Assembly datasets with different random seeds. Differing random seeds impact the network initialization and sampling of batch indices for our training pipeline. We report the results of this experiment in Table (mean, std. err.) for the MoF metric (cf. Tab.~\ref{tab:reproducibility}).

\begin{table}
\begin{tabular}{ | c | c | c | c | }
 \hline
 Dataset & 50 Salads (Mid) & 50 Salads (Eval) & Desktop Ass. \\ 
 \hline
 MoF (\%) & (43.0, 1.9) & (56.4, 2.8) & (62.9, 2.8) \\ 
 \hline
\end{tabular}
\caption{MoF results for five runs recorded as (mean, std. err.).}
\label{tab:reproducibility}
\end{table}


\end{document}